\documentclass[10pt,twocolumn,letterpaper]{article}

\usepackage{cvpr}              
\usepackage{algorithm}
\usepackage{algorithmic}
\usepackage{bbding}

\usepackage[utf8]{inputenc} 
\usepackage{booktabs}       
\usepackage{amsfonts}       
\usepackage{nicefrac}       
\usepackage{microtype}      
\usepackage{xcolor}         
\usepackage{amsmath}
\usepackage{wrapfig}
\graphicspath{{Figs/}}
\PassOptionsToPackage{prologue,dvipsnames}{xcolor}

\definecolor{cvprblue}{rgb}{0.21,0.49,0.74}
\usepackage[pagebackref,breaklinks,colorlinks,allcolors=cvprblue]{hyperref}

\def\paperID{5202} 
\def\confName{CVPR}
\def\confYear{2025}

\title{LeanGaussian: Breaking Pixel or Point Cloud Correspondence \\ in Modeling 3D Gaussians}

\makeatletter
\renewcommand{\@fnsymbol}[1]{%
     \ifcase#1\or *\or \#\or ***\or ****\or *****\else\@ctrerr\fi%
   }
\makeatother

\author{Jiamin Wu$^{1,2}\footnotemark[1]$ \quad Kenkun Liu$^{3}\footnotemark[1]$ \quad Han Gao$^{1}$ \quad Xiaoke Jiang$^{1}\footnotemark[2]$ \quad Yuan Yao$^{2}$ \quad Lei Zhang$^{1}$ \\
{\normalsize $^1$International Digital Economy Academy (IDEA)}
\\{\normalsize $^2$Hong Kong University of Science and Technology
}
\\ {\normalsize $^3$The Chinese University of Hong Kong, Shenzhen} 
}

\begin{document}
\maketitle

\footnotetext[1]{Equal contribution. Work done during an internship at IDEA.}
\footnotetext[2]{Corresponding author.}
\begin{abstract}
Recently, Gaussian splatting has demonstrated significant success in novel view synthesis. Current methods often regress Gaussians with pixel or point cloud correspondence, linking each Gaussian with a pixel or a 3D point. This leads to the redundancy of Gaussians being used to overfit the correspondence rather than the objects represented by the 3D Gaussians themselves, consequently wasting resources and lacking accurate geometries or textures.
In this paper, we introduce LeanGaussian, a novel approach that treats each query in deformable Transformer as one 3D Gaussian ellipsoid, breaking the pixel or point cloud correspondence constraints. We leverage deformable decoder to iteratively refine the Gaussians layer-by-layer with the image features as keys and values.
Notably, the center of each 3D Gaussian is defined as 3D reference points, which are then projected onto the image for deformable attention in 2D space.
On both the ShapeNet SRN dataset (category level) and the Google Scanned Objects dataset (open-category level, trained with the Objaverse dataset), our approach, outperforms prior methods by approximately 6.1\%, achieving a PSNR of 25.44 and 22.36, respectively. Additionally, our method achieves a 3D reconstruction speed of 7.2 FPS and rendering speed 500 FPS. Codes are available at \url{https://github.com/jwubz123/LeanGaussian}.
\\
\end{abstract}
\section{Introduction}
\begin{figure*}[htp]
    \centering
    \includegraphics[width=0.8\textwidth]{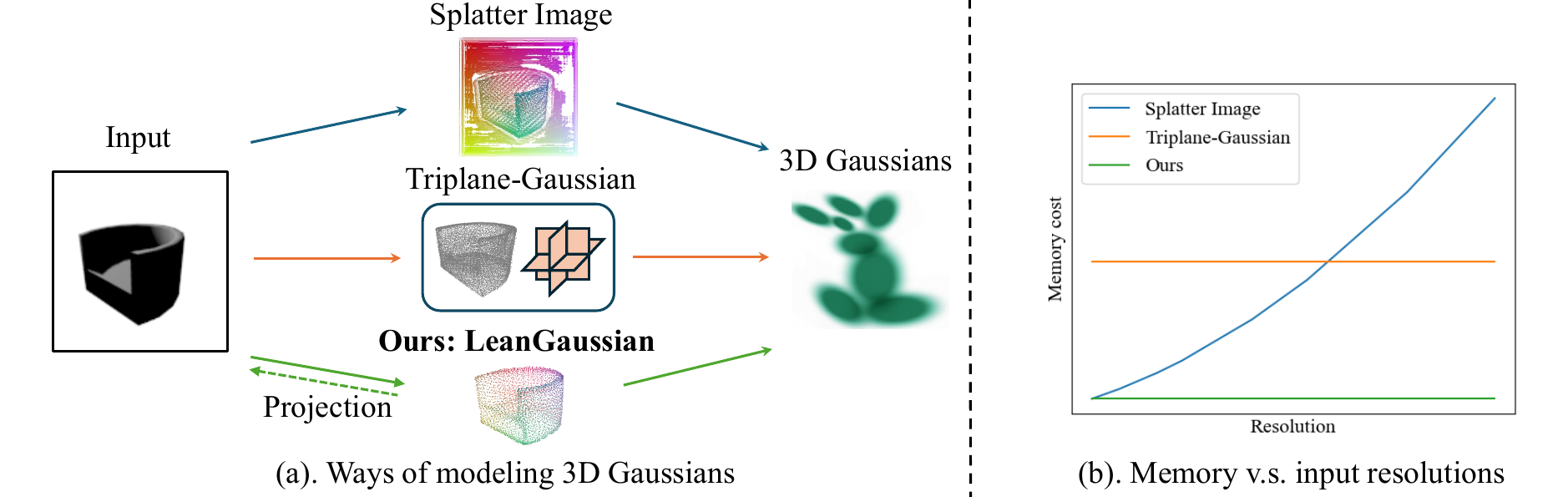}
    \caption{Comparison between our method and previous approaches. (a)Splatter Image \citep{SplatterImage} directly models Gaussians for each pixel from image features, often resulting in an overabundance of background Gaussians. Triplane-Gaussian \citep{triplane-gs} models 3D Gaussians using dense point cloud and triplane methods, leading to intricate and inefficient modeling and computation. In contrast, our method utilizes 3D Gaussians through queries, maintaining a streamlined representation. The centers of these Gaussian ellipsoids are depicted as a point cloud and shown for visualization and the dashed line means projection from 3D Gaussian to 2D plane. (b) Comparison of memory usage between our model and previous work. As resolution increases, memory costs increase quadratically for Splatter Image but not for our approach. Triplane-Gaussian has a much larger memory cost.}
    \label{fig:introduction}
\end{figure*}
Novel view synthesis (NVS) from a single view at the object level remains an unsolved computer vision problem, as evidenced by recent work on this task \citep{pixelnerf, visionnerf, SplatterImage}.
Recently, 3D Gaussian splatting (3D GS) \citep{3d-gs} has emerged as a efficient 3D representation that attains fast rendering speeds through the use of explicit 3D Gaussian ellipsoids. Generation-based models \citep{dreamgaussian, gaussianeditor} often face issues of slow speed and distortion, making them more appropriate for entertainment scenarios. For  non-generation single-forward methods, recent research efforts \citep{SplatterImage, a_pixel_more_Gau, pixelsplat, triplane-gs, instantmesh, lgm, xu2024grm, TripoSR} have attained fast rendering speeds with a reliance solely on 2D supervision, yielding outstanding outcomes.

Mainstream 3D Gaussian modeling methods falls into two primary categories as shown in \cref{fig:introduction}: the first involves direct feature lifting from individual pixels to 3D space \citep{SplatterImage, a_pixel_more_Gau, pixelsplat, lgm, xu2024grm, TripoSR}, where each pixel is associated with one Gaussian ellipsoid; the second method utilizes point clouds and triplane features \citep{triplane-gs, instantmesh}, treating each Gaussian as a point alongside its corresponding triplane feature. However, both representations require alignment with other forms, be it 2D pixels or 3D point clouds. This requirement leads to redundant Gaussians being utilized to essentially overfit other representations rather than focusing on the 3D Gaussians themselves. This clashes with the Gaussian splatting process since the rendered pixels by Gaussian splatting has no one-to-one correspondence to the 3D Gaussians. As demonstrated in \cref{fig:introduction}, Splatter Image \citep{SplatterImage}, the foremost pixel-aligned method, heavily concentrates (over half) 3D Gaussians in a single plane (the background). Triplane-Gaussian \citep{triplane-gs}, the most influenced method combining point clouds with triplane features, complicates the modeling process by regressing dense point clouds (\cref{fig:introduction} (a)).
Furthermore, the triplane representation compresses 3D space, leading to a lack of detailed information in the 3D structure and imposing a rigid grid alignment that limits flexibility \citep{lgm, qi2017pointnet}. Additionally, as illustrated in \cref{fig:introduction} (b), the computational complexity of Splatter Image increases quadratically with the input resolution due to pixel alignment regression, while Triplane-Gaussian has significantly higher memory costs because of its dense point cloud and complex Transformer design across triplane. 

To address the overfitting resulting from pixel or point cloud correspondence, we present LeanGaussian, a novel approach that breaks free from the correspondence constraint by directly modeling 3D Gaussians.
Inspired by the effective and efficient Deformable-DETR (Detection Transformer) framework in detection tasks \citep{DETR, detr0, dino, maskdino, GroundingDino, taptr, dfa3d}, where `queries' are utilized to iteratively refine the regression of bounding box (bbox) parameters (position, size and category), we leverage queries to regress and refine the 3D Gaussian parameters following the DETR paradigm, using image features as keys and values.
In our approach, each query represents a Gaussian ellipsoid with a linear transformation. Queries representing all 3D Gaussians are derived through a layer-by-layer refinement manner within the Transformer decoder. The flexibility to adjust the number of queries arbitrarily ensures that the modeling approach avoids overfitting to alternative representations, whether in pixel or point cloud form.
There is a challenge in the deformable cross-attention, a fundamental component in the deformable DETR decoder \citep{DETR}. In the original DETR detection task, the 2D `reference points,' a collection of 2D coordinates within the image feature plane, directly correspond to the bbox position, which is the regression target. These reference points play a crucial role, aiding the deformable cross-attention in pinpointing the location of the most relevant image feature. However, in our 3D context, such explicit reference points are absent. To address this issue, we define the center of each 3D Gaussian as a 3D reference point, projecting it onto the image plane under the pinhole camera model assumption and treating these projected centers as 2D reference points. These projections act as anchors on the image, guiding the selection of sampling points to extract the most relevant image features. Subsequently, these features are employed as keys and values within the deformable cross-attention mechanisms. This 3D reference point design efficiently identifies the most relevant 2D image features to each 3D Gaussian ellipsoids automatically and models lean 3D Gaussians directly from 2D images.

Our contributions can be summarized as follows:

\begin{itemize}
\item We introduce LeanGaussian, a novel and efficient method that breaks the pixel or point cloud correspondence for generating new views from a single RGB image and achieves impressive reconstruction speeds of 7.2 FPS and rendering speeds of 500 FPS.
\item We introduce a novel design of 3D reference points and project it onto the image plane to extract the most relevant image features for each 3D Gaussian ellipsoid, effectively shaping the deformable cross-attention mechanism to derive 3D Gaussians from 2D input.
\item We conducted extensive experiments to demonstrate the efficacy of our our method, achieving superior performance compared to existing methods on both category level dataset ShapeNet SRN \citep{ShapeNet-SRN} and open-category level datasets Objaverse, GSO \citep{gso}.
\end{itemize}

\section{Related work}

\paragraph{Novel view synthesis from single-view image}
Novel view synthesis (NVS) from a single RGB image is ill-posed problem in computer vision, addressed by deep learning models that regress various 3D representations. Recently, Neural Radiance Field (NeRF) transforms NVS by employing neural networks to model and render 3D objects \citep{nerf, pixelnerf, bakingnerf,nerf,nerf++,refnerf,surfacenerf,meshnerf,nerfsdf,mipnerf,Plenoxels, fastnerf, steernerf,instantnerf,LNerfAccES}. 
However, the time-consuming nature of the implicit Neural Radiance Fields (NeRF) representation and the intensive sampling algorithm utilized in volume rendering persist as challenges in the entire pipeline. 3D Gaussian splatting (3D GS) stands out as a swift alternative to NeRF owing to its explicit representation.
While generation-based approaches \citep{dreamgaussian, gaussianeditor, lgm} use diffusion to fill in missing parts, surpasses previous 3D reconstruction and NVS methods \citep{zero123, one2345} while significantly improving rendering speed. However, they suffer from long inference times and distortion, making them less suitable for certain industrial scenarios.  
Other approaches such as Wonder3D \citep{wonder3d}, InstantMesh \citep{instantmesh}, and LGM \citep{lgm} serve as multi-view models capable of generating multiple views from a single image. However, they encounter challenges related to distortion and quality dependence on the multi-view diffusion model.
In contrast, end-to-end models like LRM \citep{openlrm} and Splatter Image \citep{SplatterImage} offer less inference time ($<$ 1s) without relying on pretrained methods, although their performance may be lacking. \cite{a_pixel_more_Gau} attempts to mitigate this by regressing multiple Gaussians for each pixel in \cite{SplatterImage}, the challenge of per-pixel alignment persists.
Gamba \citep{gamba} modeling of 3D Gaussians and image tokens to a sequence, without establishing the relationship between images and 3D Gaussians, resulting in suboptimal outcomes. 
\paragraph{Deformable Transformer}
The Detection Transformer has gained significant popularity in detection tasks \citep{detr0}. Subsequent works, such as Deformable DETR \citep{DETR, dab-detr, detrx, detr-matching, dn-detr, lite-detr, taptr}, introduced deformable attention mechanisms to handle object deformations, while Cascade DETR employed a cascaded architecture for progressive refinement. Furthermore, \cite{GroundingDino} extended the structure to open-set detection by incorporating text embeddings, and \cite{taptr} applied it to point tracing in video tasks. These follow-up works have expanded the applications of deformable Transformer beyond detection. DFA3D \citep{dfa3d} and BEVFormer \citep{bevformer} are introduced to address the feature-lifting challenge in 3D detection and autonomous driving tasks. They estimate depth to convert 2D feature maps to 3D and sampling around the center of 3D bbox. In our scenario, the number of reference points far exceeds those in 3D detection, resulting in a significant computational burden when projecting 3D sampling points.

\section{Methods}
\begin{figure*}[htp]
    \centering
    \includegraphics[width=0.85\textwidth]{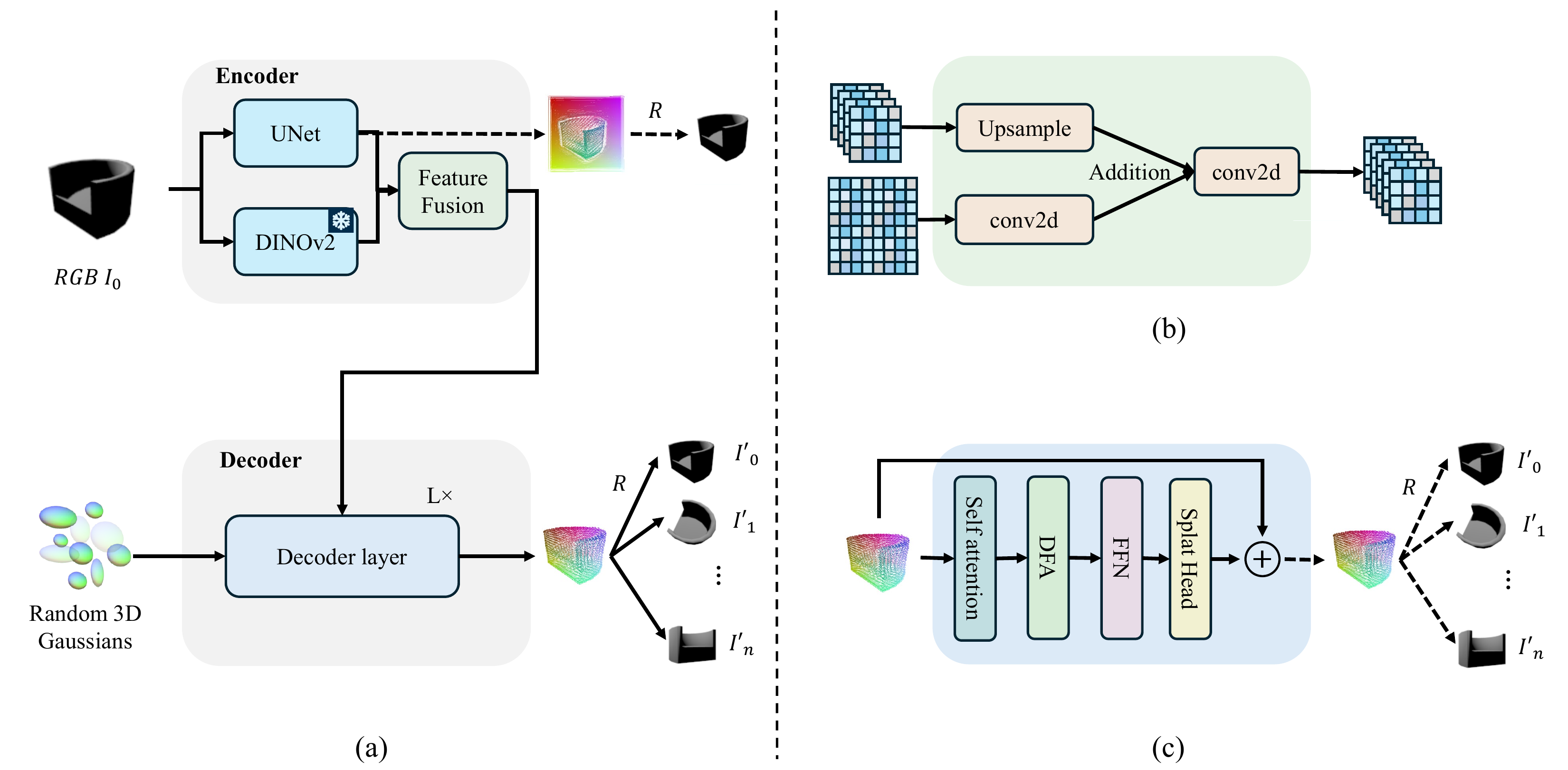}
    \caption{(a) Overview of LeanGaussian. The initial Gaussians are calculated from random queries using the splatter head. $\dashrightarrow$: steps are utilized in training only. (b) Detailed structure for feature fusion in the feature extractor. (c) Detailed structure for one decoder layer. Queries are updated at each layer and serve as input for the next layer, while the reference points are updated based on the new centers of the Gaussians and projected onto the image feature plane. GaussianDFA: deformable cross attention layer; FFN: Feed Forward Network; $\bigoplus$: updation of 3D Gaussian. $q^l$ denotes the query of $l$-th layer.}
    \label{fig:overview}
\end{figure*}

In this section, we introduce our proposed LeanGaussian for novel view synthesis (NVS) from single-view images. We begin by providing the necessary preliminaries for 3D Gaussian splatting, as detailed in \cref{sec: 3d-gs}. Next, we give the formulation of the novel view synthesis (NVS) tasks using 3D Gaussian splatting in \cref{sec: problem}. 
In \cref{sec: overview}, we describe the overall framework of our method and give the detailed design of updating 3D Gaussians with deformable Transformer. Finally, we discuss the training objective in \cref{sec: loss}. The high level difference between our model and Splatter Image is in the \cref{app: method_compare}.

\subsection{Prelimenaries} \label{sec: 3d-gs}

3D objects are represented using 3D Gaussian ellipsoids, characterized by their geometric and appearance parameters. The position and shape of an ellipsoid are determined by the mean vector $\boldsymbol{\mu}$ and the covariance matrix $\boldsymbol{\sigma}$ of its corresponding 3D Gaussian distribution \citep{EWASplatting, 3d-gs}.
Directly optimizing the covariance matrix can result
in a non-positive semi-definite matrix \citep{3d_gs_survey}. Therefore, the covariance matrix can be optimized through a combination of rotation and scaling for each ellipsoid as $\boldsymbol{\sigma} = \mathbf{R}\mathbf{S}\mathbf{S}^T\mathbf{R}^T$, where $\mathbf{R}$ represents the rotation by quaternion and $\mathbf{S}$ contains the scales in three directions (xyz) \citep{3d-gs}.

In addition to position and shape, 3D Gaussians are also parameterized by opacity $\boldsymbol{\sigma}$, which represents the probability of a light ray being blocked by the ellipsoid \citep{nerf, 3d-gs}.
As 3D Gaussian splatting \citep{3d-gs, 3d_gs_survey} can be considered an extension of point-based rendering methods \citep{3d-gs, 3d_gs_survey}, the color $C(\mathbf{p})$ of a pixel $\mathbf{p}$ on the image plane can be rendered utilizing point-based rendering techniques by the following equation:

\begin{align} \label{eq: 3d-gs-render}
    C(\mathbf{p}) & = \sum_{n = 1}^{\mathcal{N}}\mathbf{c}_n\boldsymbol{\alpha}_n\prod_{j=1}^{n-1}(1-\boldsymbol{\alpha}_j), \\
    \boldsymbol{\alpha}_n & = \boldsymbol{\sigma}_ne^{\frac{1}{2}(\mathbf{p}-\boldsymbol{\mu}_n)^T\boldsymbol{\sigma}_n^{-1}(\mathbf{p}-\boldsymbol{\mu}_n)}
\end{align}

Here $\mathcal{N}$ represents the number of Gaussians on the ray passing through the pixel $\mathbf{p}$. $\mathbf{c}_n, \boldsymbol{\sigma}_n,\boldsymbol{\mu}_n, \boldsymbol{\sigma}_n$ denotes the color, opacity, center, and covariance matrix of the $n$-th Gaussian. When it comes to color representation in the Gaussian splatting, Spherical Harmonics (SH) \citep{3d-gs} are employed to represent the view-dependent appearance of each ellipsoid. Following \cite{SplatterImage}, we use a third-order SH to represent the color, which requiring 12 parameters.

Empirically, we find that directly regressing the center of 3D Gaussians can lead to convergence difficulties. Inspired by \cite{SplatterImage}, the center coordinates $x, y, z$ are parameterized by the depth $d$ and offset values $(\Delta_{x}, \Delta_{y}, \Delta_{z})$. The depth $d$ represents the length of a ray originating from the camera center. Then, the center coordinates $\boldsymbol{\mu}$ of a 3D Gaussian can be represented by 4 parameters $(d, \Delta_{x}, \Delta_{y}, \Delta_{z})$ and acquired with the below equation,

\begin{equation} \label{eq: gaussian_center}
    \boldsymbol{\mu} = 
    \begin{bmatrix}
        x\\
        y \\
        z \\
    \end{bmatrix}
    =
    \begin{bmatrix}
        u_1d + \Delta_{x}\\
        u_2d + \Delta_{y} \\
        d + \Delta_{z}
    \end{bmatrix}
\end{equation}

\subsection{Problem formulation} \label{sec: problem}

The problem of single view NVS is formulated by learning a mapping from single view images to 3D Gaussian parameters and then render to novel views. Specifically, given a set of multi-view RGB images denoted as $\mathcal{D} = \{I_{oj}, \pi_{oj} \mid o=1, ..., O; j=1, ..., J_o\}$, where $I_{oj} \in \mathbb{R}^{H\times W \times 3}$ represents the $j$-th image of the $o$-th object with height $H$ and width $W$, and $\pi_{oj}$ represents the camera pose for image $I_{oj}$. $O, J_o$ denote the number of object and number of views for the object in the set. We aim to learn a mapping $\mathcal{G}_{\Phi}$ that maps any given input image $I_{oj}$ to a 3D object $\mathbf{G}_o$ represented by $N$ 3D Gaussians. In other words, we want to obtain $\mathbf{G}_o = \mathcal{G}_{\Phi}(I_{oj})$, where $\mathbf{G}_o = \{\boldsymbol{\mu}_{on}, \boldsymbol{\sigma}_{on}, \mathbf{S}_{on}, \mathbf{R}_{on}, \mathrm{\textbf{SH}}_{on} | n = 1, ..., N\}$. The parameters for the $n$-th Gaussian of the $o$-th object contains center $\boldsymbol{\mu}_{on}$, opacity $\boldsymbol{\sigma}_{on}$, rotation $\mathbf{R}_{on}$, scale $\mathbf{S}_{on}$, and spherical harmonic coefficients $\mathrm{\textbf{SH}}_{on}$. 
Notably, instead of directly regressing the 3D coordinates of the center $\boldsymbol{\mu}$, we regress the depth $d$ and offset values $(\Delta_{x}, \Delta_{y}, \Delta_{z})$. This choice simplifies the regression process and aids convergence. The Gaussian shape parameter $\boldsymbol{\Sigma}$ can be regressed using the scale $\mathbf{S}$ and rotation $\mathbf{R}$. 
Subsequently, given the camera parameters $\pi_{oj'}$, we can render the $j'$-th novel view $I_{oj'_{pred}} = \mathcal{R}(\mathcal{G}_{\Phi}^l(I_{oj}), \pi_{oj'})$ and supervise it with the ground-truth $I_{oj'}$. For simplicity, we focus on outlining the training and inference procedure for a single image, omitting the subscript for subsequent sections.

\subsection{LeanGaussian} \label{sec: overview}

\paragraph{Feature extractor}
Novel view synthesis from a single view is a challenging task even we construct an explict 3D representation in the middle due to depth ambiguity. To address this issue, our model incorporates a depth-aware image feature extractor that extracts both pixel-aligned features and depth-aware features from the input image. In \cref{fig:overview}(a), we utilize both UNet \citep{unet} and DINOv2 \citep{dinov2} to extract pixel-aligned and depth-aware features, respectively.
UNet is employed to leverage per-pixel correspondence in the input view and ensure consistency in the position of novel view pixels. To achieve this, we introduce a loss on the reconstructed input view. On the other hand, DINOv2 is a pretrained model on the depth prediction task, also providing structure and texture of the salient content in images \citep{openlrm}. 

In order to harness the advantages offered by the image features from both DINOv2 and UNet, a feature fusion block shown in \cref{fig:overview}(b) is designed inspired by the Feature Pyramid Networks (FPN) \citep{FPN}. The upsampled DINOv2 features is added to UNet features followed with a convolution layer. The aggregated features are further processed using another convolution layer. To reduce computational costs, the resulting features with the original resolution are downsampled using a 2D convolution network with a large stride and kernel size (14 in our setting). The fused feature $\mathbf{F}$ serves as the keys and values in the deformable cross-attention within the Transformer decoder layers.

\paragraph{3D Gaussian initialization} \label{sec: init}

Our model necessitates initialized 3D Gaussians for projection, yet random initialization can lead to convergence challenges. To address this, we initialize queries randomly and employ a splatting head $\mathbf{G}_{\mathrm{init}} = \mathcal{S}(\mathbf{q}_{\mathrm{init}})$ to regress the Gaussian positions. Our model initializes a convolution layer with weights through Xavier uniform initialization, while biases are set using constant values predetermined as hyperparameters to control the range of Gaussian parameters.

\paragraph{3D Reference point projection in GaussianDFA} \label{sec: ref_points}

\begin{wrapfigure}{l}{0.3\textwidth}
    \centering
    \includegraphics[width=0.3\textwidth]{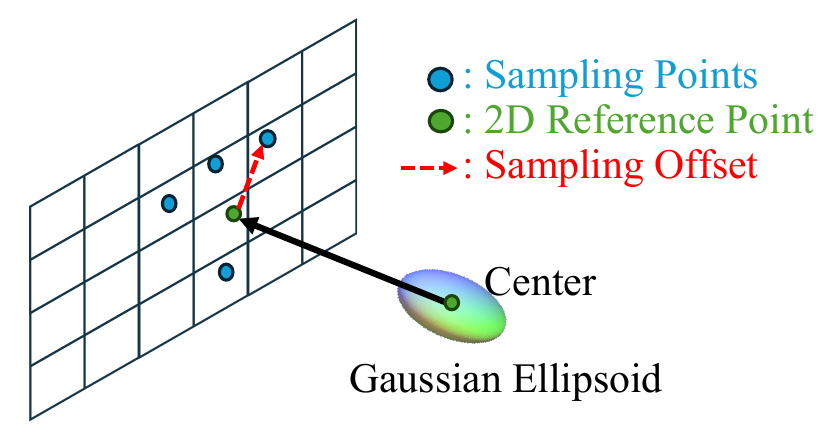}
    \caption{3D Gaussians' centers are projected onto the image feature maps. By training sampling offsets, deformable attention is performed on the features at the sampling points and queries.}
    \label{fig:detail_decoder}
\end{wrapfigure}

Deformable cross attention is a prevalent technique in detection scenarios, utilizing reference points such as the center of bounding boxes to extract features on the image plane for cross-attention computations. However, our scenario diverges as our Gaussian ellipsoids exist in 3D space, not on the image plane, hence lacking direct reference points. In contrast to the approach in \cite{dfa3d} for 3D detection tasks, where a single reference point in 3D is chosen and 3D points around the bounding box center are sampled, we designate the centers of Gaussians as our 3D reference points. Given the multiple 3D reference points, sampling around them and subsequently projecting these points incurs significant computational costs. Therefore, in our design, we project the center of a 3D Gaussian onto the image features and considering it as the reference point in 2D, as depicted in Figure \cref{fig:detail_decoder}(b). By utilizing the pinhole camera model, the projection of the 3D Gaussians at the $l$-th layer in camera coordinates can be described as $\mathbf{P}^l = C_{intrin} \boldsymbol{\mu}^{l} / d^l$, where $C_{intrin}$ represents the camera's intrinsic parameters while $\boldsymbol{\mu}^l$ and $d^l$ are the center and the depth of Gaussians in the $l$-th layer. Subsequently, we train sampling offsets $\mathbf{\Delta S}^{l} = \mathrm{MLP}^{S}(\mathbf{q}^{l})$ to determine the sampling points $\mathbf{S}^{l} = \mathbf{P}^l + \mathbf{\Delta S}^{l}$ on the image features. The features utilized in the cross attention are indexed by the sampling points.

\paragraph{3D Gaussian deformable Transformer} \label{sec: decoder}
During the decoder phase, instead of directly regressing $N$ 3D Gaussian parameters from the image feature, we employ a randomized set of $N$ queries $\mathbf{q} \in \mathbb{R}^{N \times C}$ to regress 3D Gaussians $\mathbf{G} \in \mathbb{R}^{N \times K}$, where $N$ denotes the number of 3D Gaussians used to represent an object, $K$ is the dimension of 3D Gaussian parameters and $C$ signifies the hidden dimension. In each decoder layer, we iteratively update both the queries and the 3D Gaussians ($\mathbf{\Delta G} \in \mathbb{R}^{N \times K}$) to obtain the appropriate parameters for the 3D Gaussians.

As depicted in \cref{fig:overview} (c), within each layer, the queries $\mathbf{q}$ are first updated by a deformable cross attention $\mathrm{GaussianDFA}$ with the projected reference points in \cref{sec: ref_points}. 
To be more precise, the attention scores in the $l$-th layer for the sampling points $\mathbf{S}^l$ are computed using Multilayer Perceptron (MLP) on queries, while the values are obtained by bilinearly projecting the image feature at the sampling points. The update of queries cooresponding to the $n$-th Gaussian at the $l$-th layer from $\mathbf{q}_{n}^{l}$ to $\mathbf{q}_{n}^{l'}$ through deformable cross-attention is shown in Equation \cref{eq: cross_attn}. 

\begin{align} \label{eq: cross_attn}
    & \mathbf{q}_{n}^{l'} = \sum_{p=1}^{N_P}A_{p}\mathrm{Bilinear}(\mathbf{F}, \mathbf{P}^l + \mathbf{\Delta S}_{p}^l), \\
    & A_{p} = \mathrm{MLP}^{A}(\mathbf{q}_{n}^l)
\end{align}

In this equation, $N_P$ represents the number of sampling points, and $\mathrm{MLP}^{A}$ corresponds to the learnable MLP responsible for generating the attention weights. 

Subsequently, a self-attention operation $\mathrm{SelfAttn}$ is employed to enable information exchange among all ellipsoids, followed by a feed-forward network $\mathrm{FFN}$. With $\mathbf{P}^{l}$ denoting the reference points at layer $l$ with detail in \cref{sec: ref_points}, the decoder layer can be described as \cref{eq: decoder_layer}:

\begin{equation} \label{eq: decoder_layer}
    \mathbf{q}^{l+1} = \mathrm{FFN}(\mathrm{SelfAttn}(\mathrm{GaussianDFA}(\mathbf{q}^{l}), \mathbf{P}^{l}, \mathbf{F}))
\end{equation}

Ultimately, the queries are processed through a splatter head $\mathcal{S}$ to obtain the $N \times K$ parameters for 3D Gaussians as shown as $\mathbf{\Delta G} = \mathcal{S}(\mathbf{q})$.

\paragraph{Multi-layer refinement} \label{sec: multi_layer_refine}
Taking inspiration from \cite{dino}, we employ multi-layer refinement to enhance the decoding process and facilitate iterative improvement. Firstly, we obtain the initialized 3D Gaussians $\mathbf{G}_{\mathrm{init}}$ in \cref{sec: init} and proceed to train a parameter increment $\mathbf{\Delta G}$ in \cref{sec: decoder} for each layer. We then update the 3D Gaussian parameters in a sequential manner, rendering and supervising them at each layer. 
This refinement process is denoted as $\mathbf{\Delta G}^l = \mathcal{S}^l(\mathbf{q}^l)$, where $\mathcal{S}^l$ represents the splat head of the $l$-th layer. To update the 3D Gaussian parameters from Gaussians $\mathbf{G}$ in the previous layer, we apply the following change: $\mathbf{G}^{l+1} = \mathbf{G}^{l} \bigoplus \mathbf{\Delta G}^l$. Here, $\bigoplus$ denotes the update operation of $\mathbf{G}$ by $\mathbf{\Delta G}$, involving multiplication for rotation and addition for other parameters.

Moreover, we compute the refined version of the reference points and utilize these updated reference points for deformable attention in the subsequent layer. By incorporating this strategy, we ensure that each layer contributes positively to refining the Gaussian representations.

\subsection{Training objective} \label{sec: loss}
Given a RGB image $I$ as input and the viewpoint change $\pi$ between the source and target cameras, we compute the discrepancy between the ground truth image $I_{gt}$ and the rendered views from the feature extractor, as well as at each output layer of decoder. The RGB loss function can be defined as \cref{eq:loss}. 

\begin{align} \label{eq:loss}
    \mathcal{L} & = \frac{1}{\left|\mathcal{D}\right|}\sum_{I_{gt} \in \mathcal{D}} (\lambda_{E}\mathcal{L}_{\textrm{MSE}}(\hat{I}^E, I_{gt}) \\
    & + \sum_{l=1}^{L}(\lambda_{D}\mathcal{L}_{\textrm{MSE}}(\hat{I}^D_l, I_{gt})  + \lambda_{\mathrm{LIPIS}}\mathcal{L}_{\mathrm{LIPIS}}(\hat{I}^D_l, I_{gt}))) \nonumber
\end{align}

Here, $\mathcal{L}_{\textrm{MSE}}$ represents the pixel-wise $l_2$ loss between rendered and ground truth image. $\hat{I}^E$ and $\hat{I}^D_l$ are the image rendered from the feature extractor and the $l$-th layer of the deformable Transformer, respectively.
$\mathcal{L}_{\mathrm{LIPIS}}$ is the LIPIS (Learned Perceptual Image Patch Similarity) loss \citep{LIPIS}. $\lambda_{E}$, $\lambda_{D}$, and $\lambda_{\mathrm{LIPIS}}$ are constant weights of the different loss.



\section{Experiments} \label{sec: exp}
We assess the effectiveness of our method in NVS using single-view RGB image in this section. The detailed experiment settings are in \cref{app: exp_settings}.
The dataset and evaluation metrics employed in our study, followed by an explanation of our experimental setup.
The qualitative and quantitative comparison of novel view synthesis between our method and the previous approaches are in \cref{sec: results}. Tthe nessaesary ablation studies for our model design are in \cref{sec: ablation}
More visualization and experimental details can be found in \cref{app: exp}.

\paragraph{Dataset and evaluation metrics}
In our study, we employ two datasets, one at the category level and the other at the open-category level. The category-level dataset, ShapeNet-SRN \cite{ShapeNet-SRN}, consists of two object classes: 'Cars' and 'Chairs'. The open-category dataset, Objaverse LVIS \citep{objaverse}, which filters the high quality data from the Objaverse with 1156 object categories commonly encountered in daily life. We evaluate the model trained on Objaverse using the Google Scanned Objects (GSO) benchmark \cite{gso}, which includes 1,030 3D objects across 17 categories. For GSO, we utilize rendered images from Free3D \cite{free3d}, following the dataset splitting methods outlined in zero-1-to-3 \cite{zero123} and Free3D \cite{free3d}. Further details about the datasets are in \cref{app: dataset}.

We assess our results using standard NVS metrics: the Peak Signal-to-Noise Ratio (PSNR), Structural Similarity (SSIM), and the Learned Perceptual Image Patch Similarity (LPIPS). Additionally, visual quality is also taken into consideration during evaluation.

\subsection{Comparison with SOTA methods} \label{sec: results}
In this section, we compare our method with existing approaches on two benchmarks: the GSO dataset for general object testing \citep{gso} and the ShapeNet SRN chair and car datasets for category-level testing \citep{ShapeNet-SRN}. We analyze the performances both qualitatively and quantitatively.

\paragraph{Quantitative comparison}
\begin{table}[htp]
  \caption{Quantitative results trained on Objaverse LVIS and tested on GSO. 3D sup. means need 3D supervision.}
  \label{tab: gso_results}
  \centering
  \footnotesize
  \begin{tabular}{lllll}
    \toprule
       Method & PSNR $\uparrow$     & SSIM $\uparrow$ & LPIPS $\downarrow$ & 3D sup. \\
    \midrule
    OpenLRM \cite{openlrm} & 18.06 & 0.840 & 0.129 & \XSolidBrush \\
    Splatter Image \cite{SplatterImage}  & 21.06 & 0.879 & 0.111 & \XSolidBrush \\
    Triplane-Gaussian \citep{triplane-gs} & 18.61 & 0.853 & 0.159 & \Checkmark \\
    \midrule
    Ours & \textbf{22.36} & \textbf{0.884} & \textbf{0.105} &\XSolidBrush \\
    \bottomrule
  \end{tabular}
\end{table}

\begin{table*}
  \caption{Quantitative results of novel view synthesis from single view on the ShapeNet-SRN dataset. Average PSNR: 25.44}
  \label{tab: novel_view_results}
  \centering
  \begin{tabular}{lllllll}
    \toprule
    \multicolumn{1}{c}{Method} & \multicolumn{3}{c}{Cars} & \multicolumn{3}{c}{Chairs}                  \\
         & PSNR $\uparrow$    & SSIM $\uparrow$ & LPIPS $\downarrow$  & PSNR $\uparrow$     & SSIM $\uparrow$ & LPIPS $\downarrow$ \\
    \midrule
    SRN \cite{ShapeNet-SRN} & 22.25  & 0.88 & 0.129 & 22.89 & 0.89 & 0.104    \\
    CodeNeRF \cite{codenerf}   &  23.80 & 0.91 & 0.128 & 23.66 & 0.90 & 0.166      \\
    ViewsetDiff w/o depth \cite{viewset_diff}    & 23.21 & 0.90 & 0.116 & 24.16 & 0.91 & 0.088  \\
    PixelNeRF \cite{pixelnerf} & 23.17 & 0.89 & 0.146 & 23.72 & 0.90 & 0.128 \\
    VisionNeRF \cite{visionnerf} & 22.88&  0.90&  0.084&  24.48&  0.92&  0.077 \\
    NeRFDiff w/o NGD \cite{nerfdiff} & 23.95 &  0.92 & 0.092 & 24.80 & 0.93 & 0.070 \\
    Splatter Image \cite{SplatterImage}  & 24.00 & 0.92 & 0.078 & 24.43 & 0.93 & 0.067 \\
    Hierarchical Splatter Image \cite{a_pixel_more_Gau} & 24.18 & 0.92 & 0.087 & 25.43 & 0.94 & 0.066 \\
    \midrule
    Ours & \textbf{25.00} & \textbf{0.93} & \textbf{0.075} & \textbf{25.87} & \textbf{0.95} & \textbf{0.065} \\
    \bottomrule
  \end{tabular}
\end{table*}

For each object, we compute the average metrics (PSNR, SSIM, and LPIPS) for the rendered images in the dataset except for the input image. Our method surpasses all the methods presented in the table.
\cref{tab: gso_results} showcase the results of training on the Objaverse LVIS dataset \citep{objaverse} and test on the GSO dataset \citep{gso}, our model outperforms previous methods. Particularly, our model achieves a higher PSNR score of 22.36, surpassing previous methods by 1.28. Moreover, Triplane-Gaussian requires 3D supervision while other methods do not. On another note, for the testing of Triplane-Gaussian, we use the lightweight checkpoint provided training under Objaverse-LVIS dataset. In the results outlined in \cref{tab: novel_view_results} on the ShapeNet SRN dataset, our model also achieves State-of-the-Art (SoTA) results in both the chairs and cars categories.

\begin{figure}[hpt]
    \centering
    \includegraphics[width=0.48\textwidth]{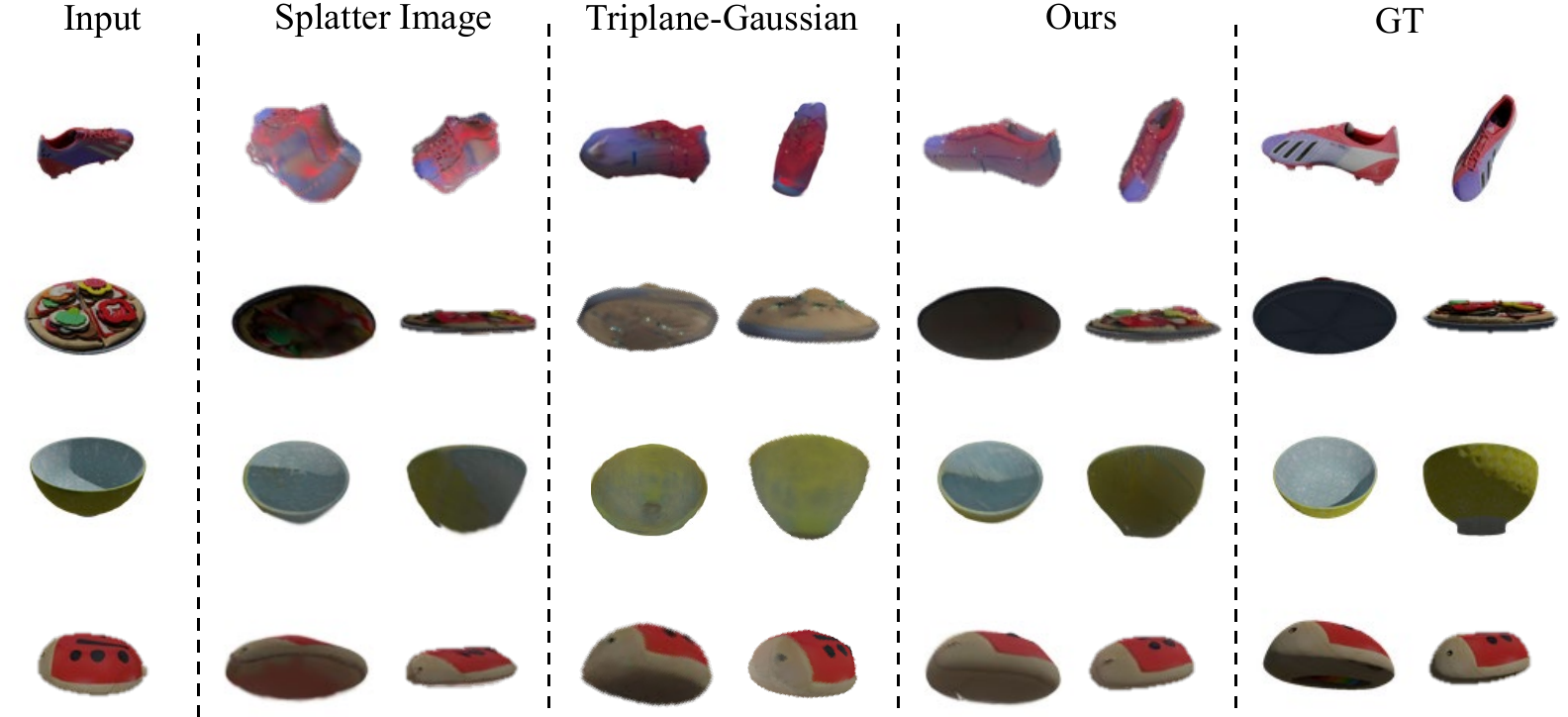}
    \caption{Comparison of NVS results between models trained on Objaverse LVIS and tested on GSO dataset.}
    \label{fig: gso_visualization}
\end{figure}

\paragraph{Qualitive comparison}

\begin{figure}[!htp]
    \centering
    \includegraphics[width=0.5\textwidth]{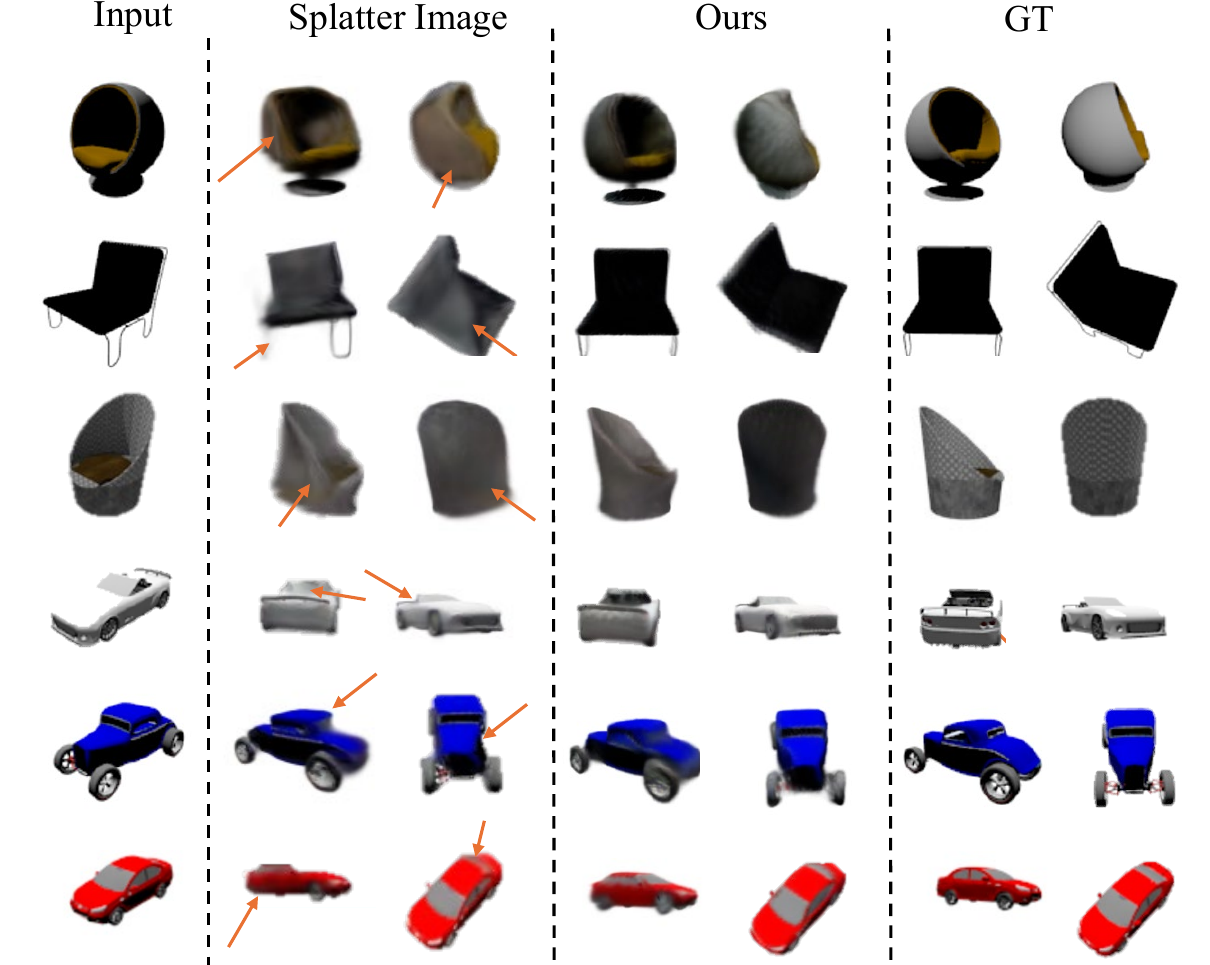}
    \caption{Comparison of NVS results between models on ShapeNet SRN. In the chairs dataset, certain objects in the Splatter Image exhibit subpar performance in terms of geometry and color accuracy. For the cars dataset, the white car shows incorrect seat color, whereas the blue and red cars exhibit incorrect geometry.}
    \label{fig: shapenet_vis}
\end{figure}

To showcase the visual quality of our approach, we present visualizations of the generated novel views on both GSO and ShapNet SRN in \cref{fig: gso_visualization} and \cref{fig: shapenet_vis}, respectively. In the Splatter Image, an additional pizza is observed at the back of the plate, and the bowl exhibits an incorrect coloration. Splatter Image frequently predicts incorrect geometries, whereas Triplane-Gaussian often exhibits inaccuracies in texture rendition. Furthermore, we provide a comparison with generation-based models such as LGM \citep{lgm} and InstantMesh \citep{instantmesh} in \cref{app: visualization} \cref{fig: generation_compare}.
For further insights, additional visualizations, including a quality comparison between Gamba \citep{gamba} and our model, can be found in \cref{app: visualization}. 
We also test the model for the in-the-wild figures and provide the visualization in \cref{fig: more_result}. Additionally, we include point cloud representations (center of Gaussians) to compare the 3D geometry of the objects in \cref{app: visualization}.



\paragraph{View-wisely comparison}

\begin{figure*}
    \centering
    \includegraphics[width=1.0\textwidth]{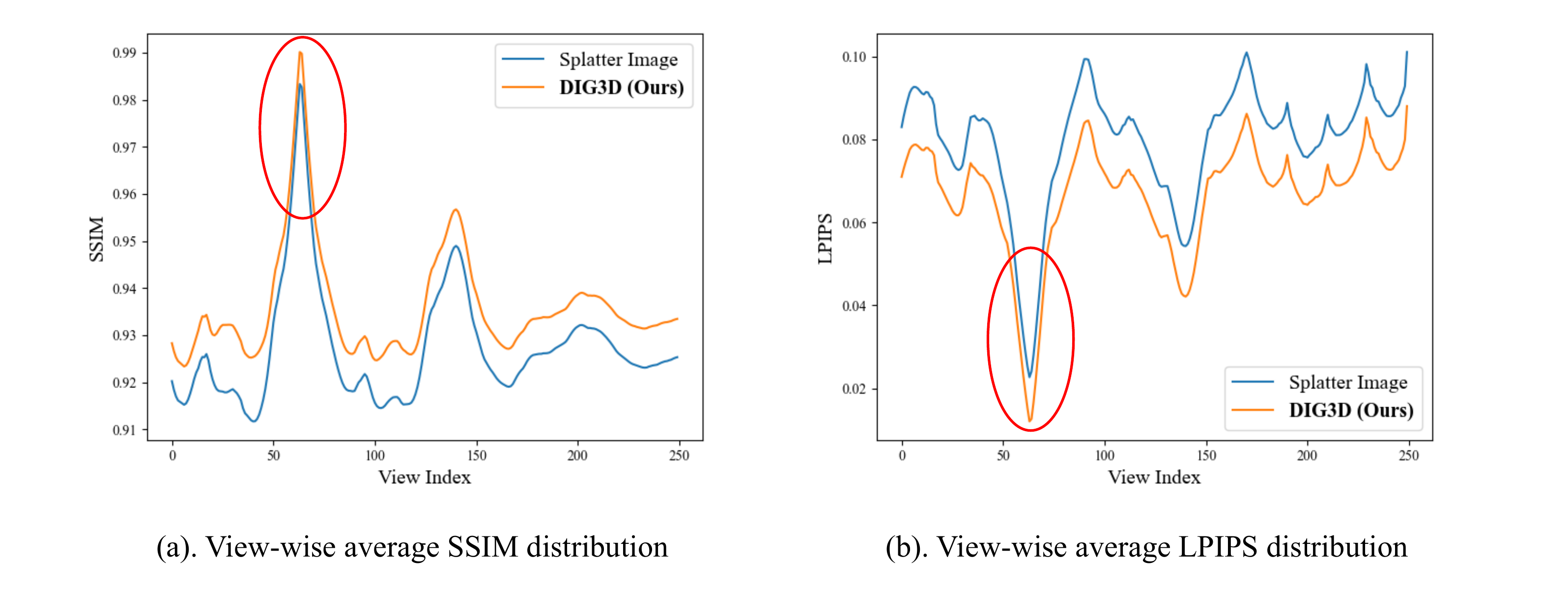}
    \caption{Comparison between our methods and previous methods on PSNR, SSIM, and LPIPS for each view in ShapeNet SRN chair dataset. The input view is at index 64, the views far away from the input view are in the shaded areas.}
    \label{fig:remove_near}
\end{figure*}

As discussed earlier, previous methods like Splatter Image and Triplane-Gaussian overfit other representation instead of 3D Gaussians themselves and may take a shortcut of the input view, yielding satisfactory results for views close to the input view but perform poorly for other views \citep{humannerf}. The results of each view are shown in \cref{fig:remove_near}. Previous methods perform similar to our method close to the input view but the difference is significant for the views far away (shaded areas). 

Moreover, We specifically choose views with an elevation or azimuth difference exceeding 45 degrees compared to the input view across all objects in the GSO dataset. Subsequently, we compute the mean results across all views for all objects in the GSO dataset. The result is shown in \cref{tab: far_view_gso}. Our model's performance remains similar (PSNR reduced by 0.35dB), whereas previous methods experience a significant decline (PSNR reduced by around 1.5dB). We also provide the result in ShapeNet SRN Chair dataset in \cref{app: remove_near}.

\begin{table}
  \caption{Novel Views for removing the near views to input view for GSO dataset}
  \small
  \label{tab: far_view_gso}
  \centering
  \begin{tabular}{llll}
    \toprule
    \multicolumn{1}{c}{Method} 
      & PSNR $\uparrow$     & SSIM $\uparrow$ & LPIPS $\downarrow$  \\
    \midrule
    Splatter Image  & 19.7071 & 0.8235 & 0.1671  \\
    Triplane-Gaussian & 17.1321 & 0.8132 & 0.1863 \\
    \midrule
    Ours & \textbf{22.0127} & \textbf{0.8727} & \textbf{0.1115}   \\
    \bottomrule
  \end{tabular}
\end{table}

\paragraph{Inference time comparison}

\begin{table}[!htp]
  \caption{Inference time. 3D: 3D Reconstruction; R: rendering. Inference: from a image to 250 novel views. Unit in second.}
  \label{tab: speed}
  \small
  \centering
  \begin{tabular}{llll}
    \toprule
    Method & 3D $\downarrow$ & R $\downarrow$ & Inference $\downarrow$     \\
    \midrule
    PixelNeRF \cite{pixelnerf} & \textbf{0.005} & 1.2200 & 304.530  \\
    OpenLRM \cite{openlrm} & 2.240 & 2.1200 & 532.240 \\
    LGM \cite{lgm} & 7.434 & 0.023 & 7.934 \\
    Splatter Image \cite{SplatterImage}  & 0.022 & 0.0025 & 0.647  \\
    Triplane Gaussian \cite{triplane-gs} & 1.281 & 0.0025 & 1.906 \\
    \midrule
    Ours & 0.140 & \textbf{0.0018} & \textbf{0.590}   \\
    \bottomrule
  \end{tabular}
\end{table}

We conducted speed tests for inference on the ShapeNet SRN dataset using a single A100 GPU.
The 3D reconstruction process takes real time for the forward pass (0.14s), our inference process is also fast ($<$ 1s) thanks to the quick rendering speed (0.002s) of Gaussian splatting (with the number of Gaussians as 10000).

We conducted speed tests for inference on a single A100 GPU. As depicted in \cref{tab: speed}, our inference process is rapid (less than 1s) due to the fast rendering speed (0.002s) of Gaussian splatting. Additionally, leveraging deformable attention, our 3D reconstruction speed (0.14s) surpasses that of other heavy Transformer-based models like the one in \cite{openlrm} (2.24s) or in \cite{triplane-gs} (1.28s). Additionally, even Triplane Gaussian here is the lightweight version, its dense Transformer and complex triplane design makes the inference slow. For the methods like LGM \citep{lgm} that first use generation models to generate multiple views from the single input view, the inference speed is much more slow.
During inference, our model executes a single forward pass to obtain the 3D Gaussian parameters and multiple passes to generate novel views. Our rendering speed is faster compared to Splatter Image because fewer Gaussians are needed to represent 3D objects. Furthermore, by utilizing a fixed number of Gaussians and disregarding image resolution, our method maintains  an almost constant rendering time and slow memory cost change as resolution increases, in contrast to previous pixel-aligned methods that suffer from quadratic growth. This comparison is detailed in \cref{fig: ablation_param}.

\subsection{Ablation study} \label{sec: ablation}
To show the significance of each component and selection of parameters of our method, we provide ablation study. To manage the computational cost, we trained the ablation models using a shorter training schedule of 100k iterations on ShapeNet SRN dataset, following Splatter Image \citep{SplatterImage}. 

\begin{table}[htp]
  \footnotesize
  \caption{Ablation study on model design.}
  \label{tab: ablation_study}
  \centering
  \begin{tabular}{llll}
    \toprule
    \multicolumn{1}{c}{Method} & \multicolumn{3}{c}{Chairs}                  \\
      & PSNR $\uparrow$     & SSIM $\uparrow$ & LPIPS $\downarrow$ \\
    \midrule
    w/o 3D reference point  & 23.30 & 0.910 & 0.105 \\
    w/o multi-layer refinement & 21.11 & 0.897 & 0.133 \\
    w/o UNet feature & 22.54 & 0.909 & 0.110  \\
    w/o supervison at the first stage & 21.99 & 0.905	& 0.121 \\
    Init. with feat. extractor Gaussians & 23.29 &	0.918 &	0.101 \\
    \midrule
    Full model & \textbf{23.45} & \textbf{0.920} & \textbf{0.100}   \\
    \bottomrule
  \end{tabular}
\end{table}

We offer a comprehensive ablation study detailing the specifics of our design in \cref{tab: ablation_study}. As depicted in the table, supervision in the initial stage enhances the PSNR by approximately 1.5. Utilizing the 3D Gaussians from the first stage as the initialization for the second stage, rather than Gaussians from random queries, yields minimal differences in the results.
However, without the deformable decoder, regressing 3D Gaussians from models like DINOv2 or UNet (which has a per-pixel alignment or per-patch (several pixels) alignment) may underperform compared to our model, as shown in \cref{tab: ablation_study_enc_dec}.

In the previous sections, we emphasized the significance of both the multi-layer refinement and 3D reference point design. Additionally, as the image extractor comprises both UNet and DinoV2, we conducted an ablation study to evaluate the impact of removing UNet. 

In \cref{tab: ablation_study}, substituting bounding boxes with Gaussians in the deformable Transformer's decoder yields poor results. Removing multi-layer refinement prevents positive contributions from each layer. Without the 3D reference point, random image feature points may lack accuracy compared to 3D Gaussian positions. Omitting the UNet feature results in sole reliance on the pre-trained model, potentially unsuitable for our task. Without the deformable Transformer, regressing 3D Gaussians from models like DINOv2 or UNet may underperform compared to our model, as shown in \cref{app: ablation_param}.
Furthermore, we also do ablation study on the hyperparameter selections and other model designs with the detailed results in \cref{app: ablation_param}.

\section{Conclusion and limitation} \label{sec: conclusion}
\paragraph{Conclusion}
In this paper, we present LeanGaussian as an efficient and effective single view NVS model. Rather than establishing a one-to-one correspondence between 3D Gaussians and alternate representations (such as pixels or point clouds), we directly model the 3D Gaussians through queries in the deformable Transformer decoder. This decoder conceptualizes 3D Gaussians as queries and utilizes image features as keys and values.
By projecting the center of each Gaussian onto the image plane in deformable cross-attention, we extract the most relavant image features to refine the parameters of the 3D Gaussians. Through subsequent layers of self-attention and feed-forward mechanisms, effective interrelations among the 3D Gaussians are captured. Our model automatically identifies associated features and iteratively updates the 3D Gaussians, overcoming challenges posed by constraints in pixel or point cloud correspondence. We assess our model through comprehensive testing on general object training datasets like Objaverse and on evaluation datasets such as GSO, as well as at a category level on ShapeNet SRN Chairs and Cars datasets, outperforming all previous methods on both datasets.
\paragraph{Limitations}
Although our method achieves good results in the single-view NVS task, it still faces certain limitations. One of the limitation is that it struggles to generate accurate textures in the background of the input view due to the absence of input image. Moreover, producing clear rendering results for self-occluded parts remains a significant challenge. These constraints present promising avenues for future research.

\clearpage
\newpage


{
    \small
    \bibliographystyle{ieeenat_fullname}
    \bibliography{main}

\begin{thebibliography}{59}
\providecommand{\natexlab}[1]{#1}
\providecommand{\url}[1]{\texttt{#1}}
\expandafter\ifx\csname urlstyle\endcsname\relax
  \providecommand{\doi}[1]{doi: #1}\else
  \providecommand{\doi}{doi: \begingroup \urlstyle{rm}\Url}\fi

\bibitem[Azinovi\'c et~al.(2022)Azinovi\'c, Martin-Brualla, Goldman, Nie{\ss}ner, and Thies]{surfacenerf}
Dejan Azinovi\'c, Ricardo Martin-Brualla, Dan~B Goldman, Matthias Nie{\ss}ner, and Justus Thies.
\newblock {Neural RGB-D Surface Reconstruction}.
\newblock In \emph{IEEE/CVF Conference on Computer Vision and Pattern Recognition (CVPR)}, 2022.

\bibitem[Barron et~al.(2021)Barron, Mildenhall, Tancik, Hedman, Martin-Brualla, and Srinivasan]{mipnerf}
Jonathan~T Barron, Ben Mildenhall, Matthew Tancik, Peter Hedman, Ricardo Martin-Brualla, and Pratul~P Srinivasan.
\newblock {Mip-NeRF: A Multiscale Representation for Anti-Aliasing Neural Radiance Fields}.
\newblock In \emph{IEEE/CVF International Conference on Computer Vision (ICCV)}, 2021.

\bibitem[Carion et~al.(2020)Carion, Massa, Synnaeve, Usunier, Kirillov, and Zagoruyko]{detr0}
Nicolas Carion, Francisco Massa, Gabriel Synnaeve, Nicolas Usunier, Alexander Kirillov, and Sergey Zagoruyko.
\newblock {End-to-end object detection with transformers}.
\newblock In \emph{European conference on computer vision (ECCV)}, 2020.

\bibitem[Charatan et~al.(2024)Charatan, Li, Tagliasacchi, and Sitzmann]{pixelsplat}
David Charatan, Sizhe~Lester Li, Andrea Tagliasacchi, and Vincent Sitzmann.
\newblock {pixelSplat: 3D Gaussian Splats from Image Pairs for Scalable Generalizable 3D Reconstruction}.
\newblock In \emph{Proceedings of the IEEE/CVF Conference on Computer Vision and Pattern Recognition (CVPR)}, pages 19457--19467, 2024.

\bibitem[Chen and Wang(2024)]{3d_gs_survey}
Guikun Chen and Wenguan Wang.
\newblock {A Survey on 3D Gaussian Splatting}.
\newblock \emph{ArXiv}, 2024.

\bibitem[Chen et~al.(2024)Chen, Chen, Zhang, Wang, Yang, Wang, Cai, Yang, Liu, and Lin]{gaussianeditor}
Yiwen Chen, Zilong Chen, Chi Zhang, Feng Wang, Xiaofeng Yang, Yikai Wang, Zhongang Cai, Lei Yang, Huaping Liu, and Guosheng Lin.
\newblock {GaussianEditor: Swift and Controllable 3D Editing with Gaussian Splatting}.
\newblock In \emph{IEEE/CVF Conference on Computer Vision and Pattern Recognition (CVPR)}, 2024.

\bibitem[Deitke et~al.(2022)Deitke, Schwenk, Salvador, Weihs, Michel, VanderBilt, Schmidt, Ehsani, Kembhavi, and Farhadi]{objaverse}
Matt Deitke, Dustin Schwenk, Jordi Salvador, Luca Weihs, Oscar Michel, Eli VanderBilt, Ludwig Schmidt, Kiana Ehsani, Aniruddha Kembhavi, and Ali Farhadi.
\newblock {Objaverse: A Universe of Annotated 3D Objects}.
\newblock \emph{arXiv preprint arXiv:2212.08051}, 2022.

\bibitem[Downs et~al.(2022)Downs, Francis, Koenig, Kinman, Hickman, Reymann, McHugh, and Vanhoucke]{gso}
Laura Downs, Anthony Francis, Nate Koenig, Brandon Kinman, Ryan Hickman, Krista Reymann, Thomas~B McHugh, and Vincent Vanhoucke.
\newblock {Google scanned objects: A high-quality dataset of 3d scanned household items}.
\newblock In \emph{2022 International Conference on Robotics and Automation (ICRA)}, pages 2553--2560. IEEE, 2022.

\bibitem[Fridovich-Keil et~al.(2022)Fridovich-Keil, Yu, Tancik, Chen, Recht, and Kanazawa]{Plenoxels}
Sara Fridovich-Keil, Alex Yu, Matthew Tancik, Qinhong Chen, Benjamin Recht, and Angjoo Kanazawa.
\newblock {Plenoxels: Radiance fields without neural networks}.
\newblock In \emph{IEEE/CVF Conference on Computer Vision and Pattern Recognition (CVPR)}, 2022.

\bibitem[Garbin et~al.(2021)Garbin, Kowalski, Johnson, Shotton, and Valentin]{fastnerf}
Stephan~J. Garbin, Marek Kowalski, Matthew Johnson, Jamie Shotton, and Julien P.~C. Valentin.
\newblock {FastNeRF: High-Fidelity Neural Rendering at 200FPS}.
\newblock In \emph{IEEE/CVF International Conference on Computer Vision (ICCV)}, 2021.

\bibitem[Gu et~al.(2023)Gu, Trevithick, Lin, Susskind, Theobalt, Liu, and Ramamoorthi]{nerfdiff}
Jiatao Gu, Alex Trevithick, Kai-En Lin, Joshua~M. Susskind, Christian Theobalt, Lingjie Liu, and Ravi Ramamoorthi.
\newblock {NerfDiff: Single-image View Synthesis with NeRF-guided Distillation from 3D-aware Diffusion}.
\newblock In \emph{International Conference on Machine Learning (ICML)}, 2023.

\bibitem[Hedman et~al.(2021)Hedman, Srinivasan, Mildenhall, Barron, and Debevec]{bakingnerf}
Peter Hedman, Pratul~P. Srinivasan, Ben Mildenhall, Jonathan~T. Barron, and Paul~E. Debevec.
\newblock {Baking Neural Radiance Fields for Real-Time View Synthesis}.
\newblock In \emph{IEEE/CVF International Conference on Computer Vision (ICCV)}, 2021.

\bibitem[Hong et~al.(2024)Hong, Zhang, Gu, Bi, Zhou, Liu, Liu, Sunkavalli, Bui, and Tan]{openlrm}
Yicong Hong, Kai Zhang, Jiuxiang Gu, Sai Bi, Yang Zhou, Difan Liu, Feng Liu, Kalyan Sunkavalli, Trung Bui, and Hao Tan.
\newblock {LRM: Large Reconstruction Model for Single Image to 3D}.
\newblock In \emph{International Conference on Learning Representations (ICLR)}, 2024.

\bibitem[Jang and Agapito(2021)]{codenerf}
Wonbong Jang and Lourdes Agapito.
\newblock {CodeNeRF: Disentangled Neural Radiance Fields for Object Categories}.
\newblock In \emph{IEEE/CVF Conference on Computer Vision and Pattern Recognition (CVPR)}, 2021.

\bibitem[Kerbl et~al.(2023)Kerbl, Kopanas, Leimkuehler, and Drettakis]{3d-gs}
Bernhard Kerbl, Georgios Kopanas, Thomas Leimkuehler, and George Drettakis.
\newblock {3D Gaussian Splatting for Real-Time Radiance Field Rendering}.
\newblock In \emph{ACM Transactions on Graphics (TOG)}, 2023.

\bibitem[Li et~al.(2022{\natexlab{a}})Li, Zhang, Liu, Guo, Ni, and Zhang]{dn-detr}
Feng Li, Hao Zhang, Shilong Liu, Jian Guo, Lionel~M Ni, and Lei Zhang.
\newblock {DN-DETR: Accelerate DETR Training by Introducing Query DeNoising}.
\newblock In \emph{IEEE/CVF Conference on Computer Vision and Pattern Recognition (CVPR)}, 2022{\natexlab{a}}.

\bibitem[Li et~al.(2023{\natexlab{a}})Li, Zeng, Liu, Zhang, Li, Zhang, and shuan Ni]{lite-detr}
Feng Li, Ailing Zeng, Siyi Liu, Hao Zhang, Hongyang Li, Lei Zhang, and Lionel~Ming shuan Ni.
\newblock {Lite DETR : An Interleaved Multi-Scale Encoder for Efficient DETR}.
\newblock In \emph{IEEE/CVF Conference on Computer Vision and Pattern Recognition (CVPR)}, 2023{\natexlab{a}}.

\bibitem[Li et~al.(2023{\natexlab{b}})Li, Zhang, Xu, Liu, Zhang, shuan Ni, and yeung Shum]{maskdino}
Feng Li, Hao Zhang, Hu-Sheng Xu, Siyi Liu, Lei Zhang, Lionel~Ming shuan Ni, and Heung yeung Shum.
\newblock {Mask DINO: Towards A Unified Transformer-based Framework for Object Detection and Segmentation}.
\newblock In \emph{IEEE/CVF Conference on Computer Vision and Pattern Recognition (CVPR)}, 2023{\natexlab{b}}.

\bibitem[Li et~al.(2023{\natexlab{c}})Li, Zhang, Zeng, Liu, Li, Ren, and Zhang]{dfa3d}
Hongyang Li, Hao Zhang, Zhaoyang Zeng, Shilong Liu, Feng Li, Tianhe Ren, and Lei Zhang.
\newblock {DFA3D: 3D Deformable Attention For 2D-to-3D Feature Lifting}.
\newblock In \emph{Proceedings of the IEEE/CVF International Conference on Computer Vision}, pages 6684--6693, 2023{\natexlab{c}}.

\bibitem[Li et~al.(2024)Li, Zhang, Liu, Zeng, Ren, Li, and Zhang]{taptr}
Hongyang Li, Hao Zhang, Shilong Liu, Zhaoyang Zeng, Tianhe Ren, Feng Li, and Lei Zhang.
\newblock {TAPTR: Tracking Any Point with Transformers as Detection}.
\newblock \emph{ArXiv}, 2024.

\bibitem[Li et~al.(2023{\natexlab{d}})Li, Gao, Tancik, and Kanazawa]{LNerfAccES}
Ruilong Li, Hang Gao, Matthew Tancik, and Angjoo Kanazawa.
\newblock {NerfAcc: Efficient Sampling Accelerates NeRFs}.
\newblock In \emph{IEEE/CVF International Conference on Computer Vision (ICCV)}, 2023{\natexlab{d}}.

\bibitem[Li et~al.(2023{\natexlab{e}})Li, Li, Wang, Liao, and Yu]{steernerf}
Sicheng Li, Hao Li, Yue Wang, Yiyi Liao, and Lu Yu.
\newblock {SteerNeRF: Accelerating NeRF Rendering via Smooth Viewpoint Trajectory}.
\newblock In \emph{IEEE/CVF Conference on Computer Vision and Pattern Recognition (CVPR)}, 2023{\natexlab{e}}.

\bibitem[Li et~al.(2022{\natexlab{b}})Li, Wang, Li, Xie, Sima, Lu, Qiao, and Dai]{bevformer}
Zhiqi Li, Wenhai Wang, Hongyang Li, Enze Xie, Chonghao Sima, Tong Lu, Yu Qiao, and Jifeng Dai.
\newblock {BEVFormer: Learning Bird’s-Eye-View Representation from Multi-Camera Images via Spatiotemporal Transformers}.
\newblock \emph{European conference on computer vision (ECCV)}, 2022{\natexlab{b}}.

\bibitem[Lin et~al.(2022)Lin, Lin, Lai, Lin, Shih, and Ramamoorthi]{visionnerf}
Kai-En Lin, Yen-Chen Lin, Wei-Sheng Lai, Tsung-Yi Lin, Yichang Shih, and Ravi Ramamoorthi.
\newblock {Vision Transformer for NeRF-Based View Synthesis from a Single Input Image}.
\newblock In \emph{2023 IEEE/CVF Winter Conference on Applications of Computer Vision (WACV)}, 2022.

\bibitem[Lin et~al.(2017)Lin, Doll{\'a}r, Girshick, He, Hariharan, and Belongie]{FPN}
Tsung-Yi Lin, Piotr Doll{\'a}r, Ross~B. Girshick, Kaiming He, Bharath Hariharan, and Serge~J. Belongie.
\newblock {Feature Pyramid Networks for Object Detection}.
\newblock In \emph{IEEE/CVF Conference on Computer Vision and Pattern Recognition (CVPR)}, 2017.

\bibitem[Liu et~al.(2024{\natexlab{a}})Liu, Jin, Zeng, Han, and Zhang]{humannerf}
Kenkun Liu, Derong Jin, Ailing Zeng, Xiaoguang Han, and Lei Zhang.
\newblock {A Comprehensive Benchmark for Neural Human Radiance Fields}.
\newblock In \emph{Conference on Neural Information Processing Systems (NIPS)}, 2024{\natexlab{a}}.

\bibitem[Liu et~al.(2024{\natexlab{b}})Liu, Xu, Jin, Chen, Varma~T, Xu, and Su]{one2345}
Minghua Liu, Chao Xu, Haian Jin, Linghao Chen, Mukund Varma~T, Zexiang Xu, and Hao Su.
\newblock {One-2-3-45: Any single image to 3d mesh in 45 seconds without per-shape optimization}.
\newblock In \emph{Conference on Neural Information Processing Systems (NIPS)}, 2024{\natexlab{b}}.

\bibitem[Liu et~al.(2023{\natexlab{a}})Liu, Wu, Hoorick, Tokmakov, Zakharov, and Vondrick]{zero123}
Ruoshi Liu, Rundi Wu, Basile~Van Hoorick, Pavel Tokmakov, Sergey Zakharov, and Carl Vondrick.
\newblock {Zero-1-to-3: Zero-shot One Image to 3D Object}.
\newblock In \emph{IEEE/CVF International Conference on Computer Vision (ICCV)}, 2023{\natexlab{a}}.

\bibitem[Liu et~al.(2022)Liu, Li, Zhang, Yang, Qi, Su, Zhu, and Zhang]{dab-detr}
Shilong Liu, Feng Li, Hao Zhang, Xiao Yang, Xianbiao Qi, Hang Su, Jun Zhu, and Lei Zhang.
\newblock {DAB}-{DETR}: Dynamic anchor boxes are better queries for {DETR}.
\newblock In \emph{International Conference on Learning Representations (ICLR)}, 2022.

\bibitem[Liu et~al.(2023{\natexlab{b}})Liu, Ren, Chen, Zeng, Zhang, Li, Li, Huang, Su, Zhu, and Zhang]{detr-matching}
Siyi Liu, Tianhe Ren, Jia-Yu Chen, Zhaoyang Zeng, Hao Zhang, Feng Li, Hongyang Li, Jun Huang, Hang Su, Jun-Juan Zhu, and Lei Zhang.
\newblock {Stable-DINO: Detection Transformer with Stable Matching}.
\newblock In \emph{IEEE/CVF International Conference on Computer Vision (ICCV)}, 2023{\natexlab{b}}.

\bibitem[Liu et~al.(2023{\natexlab{c}})Liu, Zeng, Ren, Li, Zhang, Yang, yue Li, Yang, Su, Zhu, and Zhang]{GroundingDino}
Shilong Liu, Zhaoyang Zeng, Tianhe Ren, Feng Li, Hao Zhang, Jie Yang, Chun yue Li, Jianwei Yang, Hang Su, Jun-Juan Zhu, and Lei Zhang.
\newblock {Grounding DINO: Marrying DINO with Grounded Pre-Training for Open-Set Object Detection}.
\newblock \emph{ArXiv}, 2023{\natexlab{c}}.

\bibitem[Long et~al.(2024)Long, Guo, Lin, Liu, Dou, Liu, Ma, Zhang, Habermann, Theobalt, et~al.]{wonder3d}
Xiaoxiao Long, Yuan-Chen Guo, Cheng Lin, Yuan Liu, Zhiyang Dou, Lingjie Liu, Yuexin Ma, Song-Hai Zhang, Marc Habermann, Christian Theobalt, et~al.
\newblock {Wonder3D: Single Image to 3D using Cross-Domain Diffusion}.
\newblock In \emph{Proceedings of the IEEE/CVF Conference on Computer Vision and Pattern Recognition}, pages 9970--9980, 2024.

\bibitem[Mildenhall et~al.(2020)Mildenhall, Srinivasan, Tancik, Barron, Ramamoorthi, and Ng]{nerf}
Ben Mildenhall, Pratul~P. Srinivasan, Matthew Tancik, Jonathan~T. Barron, Ravi Ramamoorthi, and Ren Ng.
\newblock {NeRF: Representing Scenes as Neural Radiance Fields for View Synthesis}.
\newblock In \emph{The European Conference on Computer Vision (ECCV)}, 2020.

\bibitem[M{\"u}ller et~al.(2022)M{\"u}ller, Evans, Schied, and Keller]{instantnerf}
Thomas M{\"u}ller, Alex Evans, Christoph Schied, and Alexander Keller.
\newblock {Instant Neural Graphics Primitives with a Multiresolution Hash Encoding}.
\newblock In \emph{ACM Transactions on Graphics (SIGGRAPH)}, 2022.

\bibitem[Oquab et~al.(2024)Oquab, Darcet, Moutakanni, Vo, Szafraniec, Khalidov, Fernandez, Haziza, Massa, El-Nouby, Assran, Ballas, Galuba, Howes, Huang, Li, Misra, Rabbat, Sharma, Synnaeve, Xu, J{\'e}gou, Mairal, Labatut, Joulin, and Bojanowski]{dinov2}
Maxime Oquab, Timoth'ee Darcet, Th{\'e}o Moutakanni, Huy~Q. Vo, Marc Szafraniec, Vasil Khalidov, Pierre Fernandez, Daniel Haziza, Francisco Massa, Alaaeldin El-Nouby, Mahmoud Assran, Nicolas Ballas, Wojciech Galuba, Russ Howes, Po-Yao~(Bernie) Huang, Shang-Wen Li, Ishan Misra, Michael~G. Rabbat, Vasu Sharma, Gabriel Synnaeve, Huijiao Xu, Herv{\'e} J{\'e}gou, Julien Mairal, Patrick Labatut, Armand Joulin, and Piotr Bojanowski.
\newblock {DINOv2: Learning Robust Visual Features without Supervision}.
\newblock In \emph{Transactions on Machine Learning Research (TMLR)}, 2024.

\bibitem[Qi et~al.(2017)Qi, Su, Mo, and Guibas]{qi2017pointnet}
Charles~R Qi, Hao Su, Kaichun Mo, and Leonidas~J Guibas.
\newblock {PointNet: Deep Learning on Point Sets for 3D Classification and Segmentation}.
\newblock In \emph{Proceedings of the IEEE conference on computer vision and pattern recognition}, pages 652--660, 2017.

\bibitem[Rakotosaona et~al.(2023)Rakotosaona, Manhardt, Arroyo, Niemeyer, Kundu, and Tombari]{meshnerf}
Marie-Julie Rakotosaona, Fabian Manhardt, Diego~Martin Arroyo, Michael Niemeyer, Abhijit Kundu, and Federico Tombari.
\newblock {NeRFMeshing: Distilling Neural Radiance Fields into Geometrically-Accurate 3D Meshes}.
\newblock In \emph{International Conference on 3D Vision (3DV)}, 2023.

\bibitem[Ren et~al.(2023)Ren, Liu, Li, Zhang, Zeng, Yang, Liao, Jia, Li, Cao, Wang, Zeng, Qi, Yuan, Yang, and Zhang]{detrx}
Tianhe Ren, Siyi Liu, Feng Li, Hao Zhang, Ailing Zeng, Jie Yang, Xingyu Liao, Ding Jia, Hongyang Li, He Cao, Jianan Wang, Zhaoyang Zeng, Xianbiao Qi, Yuhui Yuan, Jianwei Yang, and Lei Zhang.
\newblock {detrex: Benchmarking Detection Transformers}.
\newblock \emph{ArXiv}, 2023.

\bibitem[Ronneberger et~al.(2015)Ronneberger, Fischer, and Brox]{unet}
Olaf Ronneberger, Philipp Fischer, and Thomas Brox.
\newblock {U-net: Convolutional networks for biomedical image segmentation}.
\newblock In \emph{Medical image computing and computer-assisted intervention (MICCAI)}, 2015.

\bibitem[Shen and Wu(2024)]{a_pixel_more_Gau}
Jianghao Shen and Tianfu Wu.
\newblock {A Pixel Is Worth More Than One 3D Gaussians in Single-View 3D Reconstruction}.
\newblock \emph{arXiv preprint arXiv:2405.20310}, 2024.

\bibitem[Shen et~al.(2024)Shen, Yi, Wu, Zhou, Zhang, Yan, and Wang]{gamba}
Qiuhong Shen, Xuanyu Yi, Zike Wu, Pan Zhou, Hanwang Zhang, Shuicheng Yan, and Xinchao Wang.
\newblock {Gamba: Marry Gaussian Splatting with Mamba for single view 3D reconstruction}.
\newblock \emph{arXiv preprint arXiv:2403.18795}, 2024.

\bibitem[Sitzmann et~al.(2019)Sitzmann, Zollhoefer, and Wetzstein]{ShapeNet-SRN}
Vincent Sitzmann, Michael Zollhoefer, and Gordon Wetzstein.
\newblock {Scene Representation Networks: Continuous 3D-Structure-Aware Neural Scene Representations}.
\newblock In \emph{Conference on Neural Information Processing Systems (NIPS)}, 2019.

\bibitem[Szymanowicz et~al.(2023)Szymanowicz, Rupprecht, and Vedaldi]{viewset_diff}
Stanislaw Szymanowicz, Christian Rupprecht, and Andrea Vedaldi.
\newblock {Viewset Diffusion: (0-)Image-Conditioned 3D Generative Models from 2D Data}.
\newblock \emph{IEEE/CVF International Conference on Computer Vision (ICCV)}, 2023.

\bibitem[Szymanowicz et~al.(2024)Szymanowicz, Rupprecht, and Vedaldi]{SplatterImage}
Stanislaw Szymanowicz, Christian Rupprecht, and Andrea Vedaldi.
\newblock {Splatter Image: Ultra-Fast Single-View 3D Reconstruction}.
\newblock In \emph{IEEE/CVF Conference on Computer Vision and Pattern Recognition (CVPR)}, 2024.

\bibitem[Tang et~al.(2024{\natexlab{a}})Tang, Chen, Chen, Wang, Zeng, and Liu]{lgm}
Jiaxiang Tang, Zhaoxi Chen, Xiaokang Chen, Tengfei Wang, Gang Zeng, and Ziwei Liu.
\newblock {LGM: Large Multi-View Gaussian Model for High-Resolution 3D Content Creation}.
\newblock \emph{European conference on computer vision (ECCV)}, 2024{\natexlab{a}}.

\bibitem[Tang et~al.(2024{\natexlab{b}})Tang, Ren, Zhou, Liu, and Zeng]{dreamgaussian}
Jiaxiang Tang, Jiawei Ren, Hang Zhou, Ziwei Liu, and Gang Zeng.
\newblock {DreamGaussian: Generative Gaussian Splatting for Efficient 3D Content Creation}.
\newblock In \emph{International Conference on Learning Representations (ICLR)}, 2024{\natexlab{b}}.

\bibitem[Tochilkin et~al.(2024)Tochilkin, Pankratz, Liu, Huang, , Letts, Li, Liang, Laforte, Jampani, and Cao]{TripoSR}
Dmitry Tochilkin, David Pankratz, Zexiang Liu, Zixuan Huang, , Adam Letts, Yangguang Li, Ding Liang, Christian Laforte, Varun Jampani, and Yan-Pei Cao.
\newblock {TripoSR: Fast 3D Object Reconstruction from a Single Image}.
\newblock \emph{arXiv preprint arXiv:2403.02151}, 2024.

\bibitem[Verbin et~al.(2022)Verbin, Hedman, Mildenhall, Zickler, Barron, and Srinivasan]{refnerf}
Dor Verbin, Peter Hedman, Ben Mildenhall, Todd~E. Zickler, Jonathan~T. Barron, and Pratul~P. Srinivasan.
\newblock {Ref-NeRF: Structured View-Dependent Appearance for Neural Radiance Fields}.
\newblock In \emph{IEEE/CVF Conference on Computer Vision and Pattern Recognition (CVPR)}, 2022.

\bibitem[Xu et~al.(2024{\natexlab{a}})Xu, Cheng, Gao, Wang, Gao, and Shan]{instantmesh}
Jiale Xu, Weihao Cheng, Yiming Gao, Xintao Wang, Shenghua Gao, and Ying Shan.
\newblock {InstantMesh: Efficient 3D Mesh Generation from a Single Image with Sparse-view Large Reconstruction Models}.
\newblock \emph{arXiv preprint arXiv:2404.07191}, 2024{\natexlab{a}}.

\bibitem[Xu et~al.(2024{\natexlab{b}})Xu, Shi, Yifan, Chen, Yang, Peng, Shen, and Wetzstein]{xu2024grm}
Yinghao Xu, Zifan Shi, Wang Yifan, Hansheng Chen, Ceyuan Yang, Sida Peng, Yujun Shen, and Gordon Wetzstein.
\newblock {GRM: Large Gaussian Reconstruction Model for Efficient 3D Reconstruction and Generation}.
\newblock \emph{arXiv preprint arXiv:2403.14621}, 2024{\natexlab{b}}.

\bibitem[Yariv et~al.(2021)Yariv, Gu, Kasten, and Lipman]{nerfsdf}
Lior Yariv, Jiatao Gu, Yoni Kasten, and Yaron Lipman.
\newblock {Volume rendering of neural implicit surfaces}.
\newblock In \emph{Conference on Neural Information Processing Systems (NIPS)}, 2021.

\bibitem[Yu et~al.(2021)Yu, Ye, Tancik, and Kanazawa]{pixelnerf}
Alex Yu, Vickie Ye, Matthew Tancik, and Angjoo Kanazawa.
\newblock {pixelNeRF: Neural Radiance Fields from One or Few Images}.
\newblock In \emph{IEEE/CVF Conference on Computer Vision and Pattern Recognition (CVPR)}, 2021.

\bibitem[Zhang et~al.(2023)Zhang, Li, Liu, Zhang, Su, Zhu, shuan Ni, and yeung Shum]{dino}
Hao Zhang, Feng Li, Shilong Liu, Lei Zhang, Hang Su, Jun-Juan Zhu, Lionel~Ming shuan Ni, and Heung yeung Shum.
\newblock {DINO: DETR with Improved DeNoising Anchor Boxes for End-to-End Object Detection}.
\newblock In \emph{The International Conference on Learning Representations (ICLR)}, 2023.

\bibitem[Zhang et~al.(2020)Zhang, Riegler, Snavely, and Koltun]{nerf++}
Kai Zhang, Gernot Riegler, Noah Snavely, and Vladlen Koltun.
\newblock {NeRF++: Analyzing and Improving Neural Radiance Fields}.
\newblock \emph{ArXiv}, 2020.

\bibitem[Zhang et~al.(2018)Zhang, Isola, Efros, Shechtman, and Wang]{LIPIS}
Richard Zhang, Phillip Isola, Alexei~A. Efros, Eli Shechtman, and Oliver Wang.
\newblock {The Unreasonable Effectiveness of Deep Features as a Perceptual Metric}.
\newblock In \emph{IEEE/CVF Conference on Computer Vision and Pattern Recognition (CVPR)}, 2018.

\bibitem[Zheng and Vedaldi(2024)]{free3d}
Chuanxia Zheng and Andrea Vedaldi.
\newblock {Free3D: Consistent Novel View Synthesis without 3D Representation}.
\newblock In \emph{IEEE/CVF Conference on Computer Vision and Pattern Recognition (CVPR)}, 2024.

\bibitem[Zhu et~al.(2021)Zhu, Su, Lu, Li, Wang, and Dai]{DETR}
Xizhou Zhu, Weijie Su, Lewei Lu, Bin Li, Xiaogang Wang, and Jifeng Dai.
\newblock {Deformable DETR: Deformable Transformers for End-to-End Object Detection}.
\newblock In \emph{The International Conference on Learning Representations (ICLR)}, 2021.

\bibitem[Zou et~al.(2023)Zou, Yu, Guo, Li, Liang, Cao, and Zhang]{triplane-gs}
Zixin Zou, Zhipeng Yu, Yuanchen Guo, Yangguang Li, Ding Liang, Yan-Pei Cao, and Song-Hai Zhang.
\newblock {Triplane Meets Gaussian Splatting: Fast and Generalizable Single-View 3D Reconstruction with Transformers}.
\newblock \emph{ArXiv}, 2023.

\bibitem[Zwicker et~al.(2001)Zwicker, Pfister, van Baar, and Gross]{EWASplatting}
M. Zwicker, H. Pfister, J. van Baar, and M. Gross.
\newblock {EWA volume splatting}.
\newblock In \emph{IEEE Visualization (IEEE VIS)}, 2001.

\end{thebibliography}
}

\clearpage
\newpage

\appendix
\setcounter{page}{1}
\section{Appendix / supplemental material}
\subsection{Model comparison} \label{app: method_compare}
\begin{figure*}[htp]
    \centering
    \includegraphics[width=0.8\textwidth]{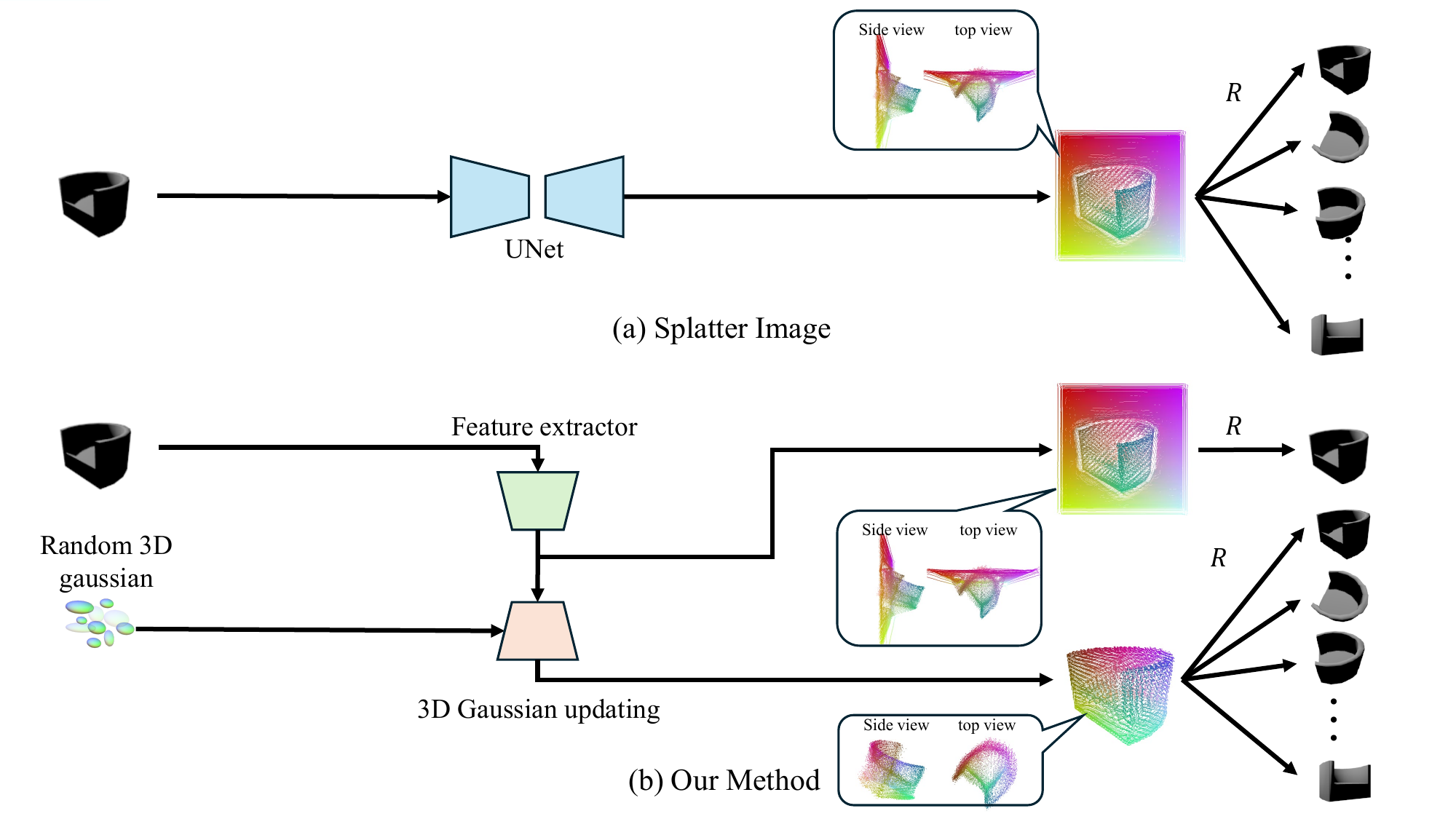}
    \caption{Comparison of our model and Splatter Image}
    \label{fig: model_compare}
\end{figure*}

We compare the stucture of Splatter Image and our method in \cref{fig: model_compare}. Instead of regressing the parameters directly from the image feature, we iterative update the random initialized 3D objects with the guidance of image feature. This design keeps a 3D and avoid the per-pixel alignment in previous method. 

\subsection{Experiment details} \label{app: exp}

\paragraph{Parameters} \label{app: param}
Similar to \cite{SplatterImage}, we utilizes 24 parameters to represent each 3D Gaussian ellipsoid. These parameters are distributed as follows: 4 parameters for $\boldsymbol{\mu}$ (1 for depth and 3 for position offset), 7 parameters for $\mathbf{\Sigma}$ (3 for scale $\mathbf{S}$ and 4 for rotation $\mathbf{R}$), 1 parameter for opacity $\sigma$, and 12 parameters for appearance $SH$. \\

For the decoder architecture, we utilized 2 layers to strike a balance between performance and computational efficiency. Regarding the representation of a 3D object, we employed 10,000 Gaussians to capture its shape and properties. 

\paragraph{Experiment settings} \label{app: exp_settings}
Our model on ShapeNet SRN was trained on a single 80G A100 GPU over a period of 2 weeks, amounting to 1 million training steps. The training process followed a similar approach as presented in \cite{SplatterImage}. During the initial 800k steps, we trained the model using only the RGB loss. Subsequently, we introduced the LPIPS loss to further enhance the training process. For the model on Objaverse, we train it on 4 A100 GPU cards for 4 days. The optimizer we utilize is Adam with learning rate 1e-5.\\

We test our model with different random seed in inference and the results are quite similar. We show the mean of 10 experiments in all quantity results.

\paragraph{Datasets} \label{app: dataset}
We train on the two categories separately to get category level 3D reconstruction models. For the ShapeNet SRN dataset, the training set includes 2458 car objects and 4612 chair objects, while the validation set has 352 car objects and 662 chair objects. The testing set comprises 704 car objects and 1317 chair objects. We follow the data splitting of ShapeNet-SRN dataset itself.
ShapeNet-SRN provides multiple-view RGB images with corresponding camera intrinsics and poses. The dataset also includes a predefined training and testing split. For the validation and testing sets, there are 250 views per object, while the training set has 50 views per object. Each image in the dataset has a resolution of $128\times128$. 

The second dataset we utilize is Objaverse LVIS \citep{objaverse}, an open-category dataset containing 1156 object categories of various objects commonly found in everyday life. To train the model, we utilize the rendered images by zero-1-to-3 \citep{zero123}.
To evaluate our model's performance on open-category scenarios, we test it on the Google Scanned Objects (GSO) dataset \citep{gso}.

\subsection{More results}


\paragraph{Quality results} \label{app: visualization}

\begin{figure}[htp]
    \centering
    \includegraphics[width=0.49\textwidth]{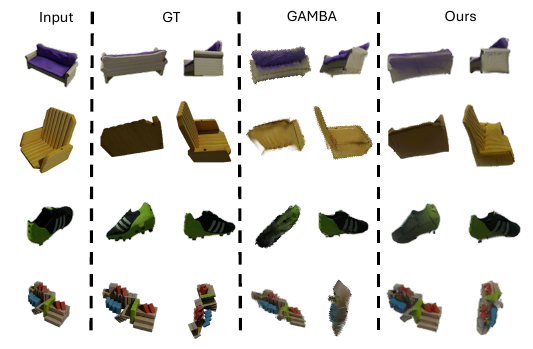}
    \caption{Novel view rendering visualization comparing to Gamba training on Objaverse and testing on GSO.}
    \label{fig: gamba}
\end{figure}

\begin{figure*}[htp]
    \centering
    \includegraphics[width=0.9\textwidth]{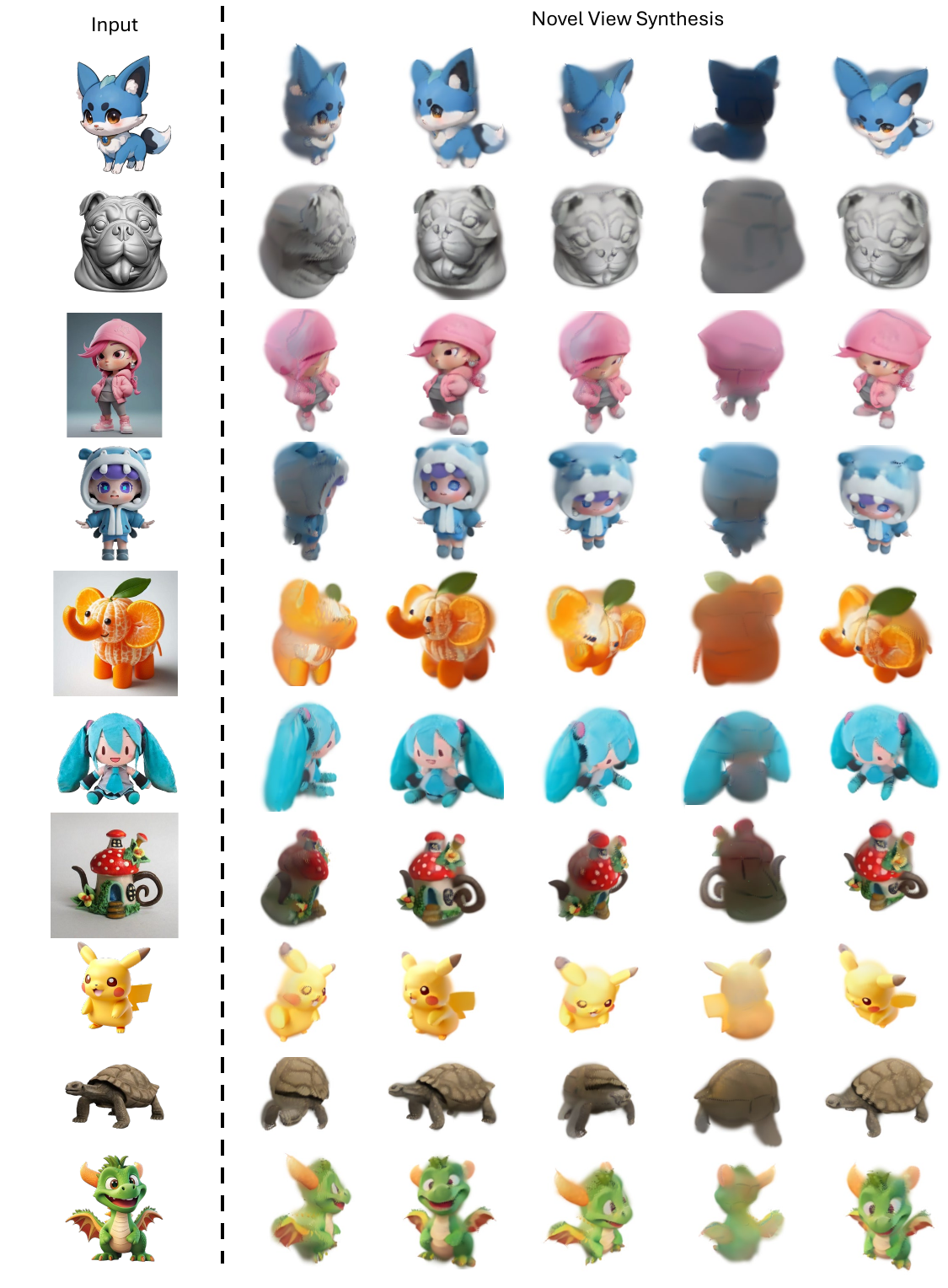}
    \caption{More qualitative results for in-the-wild images, which are widely used by image-to-3D generative methods.}
    \label{fig: more_result}
\end{figure*}

We test the new method Gamba \citep{gamba} on GSO dataset and show the comparison in \cref{fig: gamba}. We also provide more visualization on the objects widely utilized in 3D generation in \cref{fig: more_result}.

\begin{figure}[!htp]
    \centering
    \includegraphics[width=0.48\textwidth]{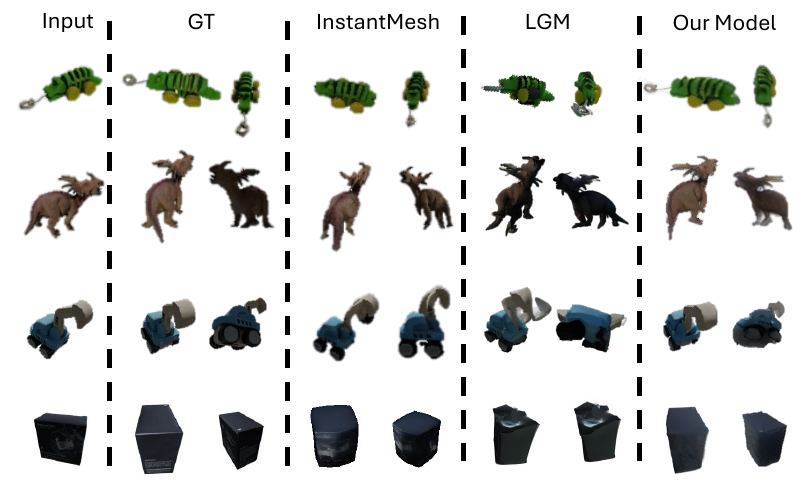}
    \caption{Comparison of NVS results between generation-based models and our model trained on Objaverse LVIS and tested on GSO shows significant differences. Both LGM and InstantMesh are trained on multi-view input and initially employ a diffusion model to generate multiple views from a single input image.}
    \label{fig: generation_compare}
\end{figure}

\begin{figure*}[t]
    \centering
    \includegraphics[width=1.0\textwidth]{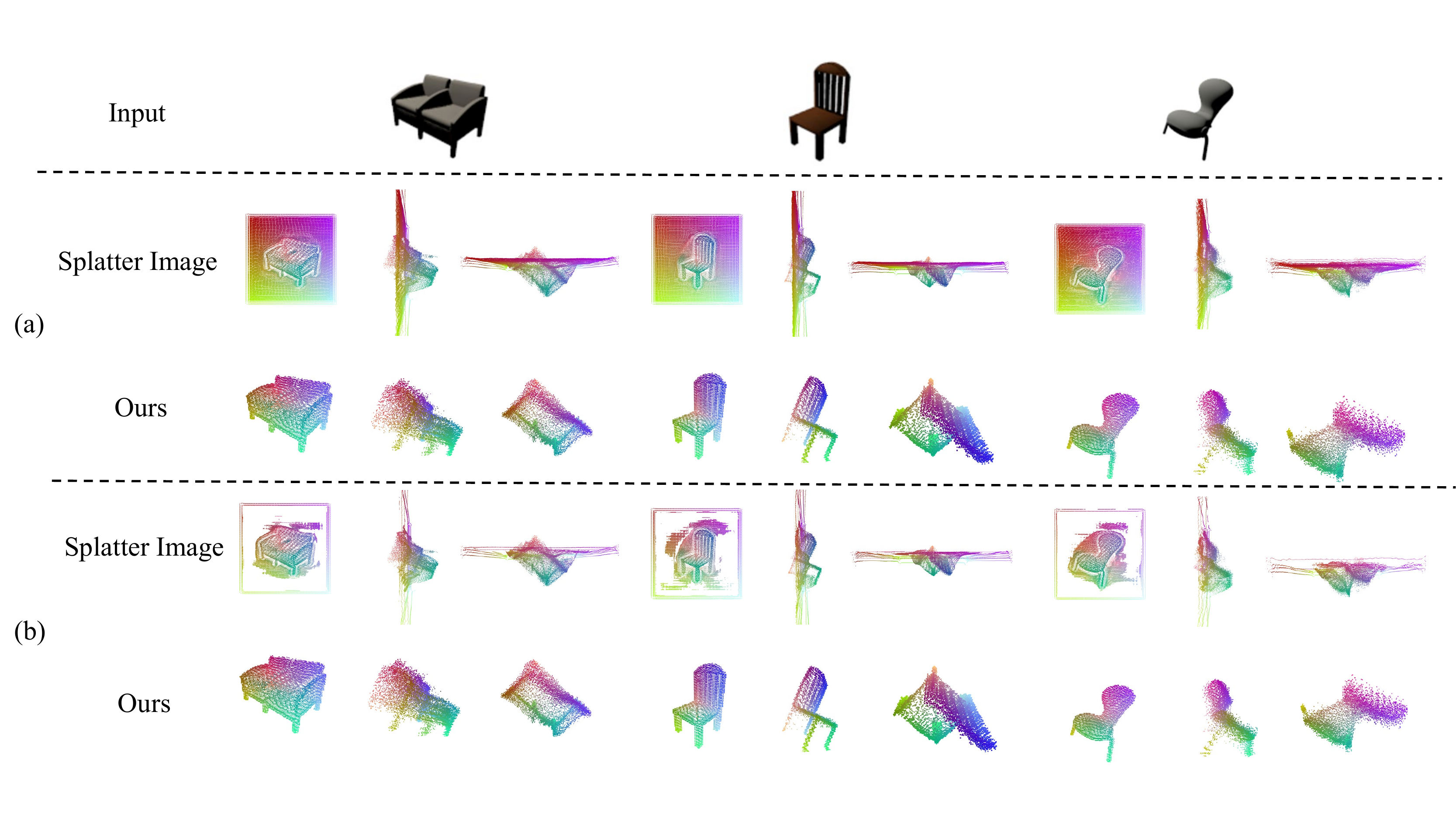}
    \caption{Point Cloud visualization. We draw the point cloud with the center of each Gaussian as a point. We show the front view, side view and top view of each point cloud. The top line of the figure shows the input view of each object. (a) shows the points without filtering. (b) shows the first 50\% points with the largest opacity.}
    \label{fig:point_cloud}
\end{figure*}

To demonstrate that our method captures a meaningful 3D structure instead of merely relying on a shortcut to the supervised views, we have visualized the point cloud with the centers of Gaussians in Figure \cref{fig:point_cloud}. In Splatter Image, each pixel in the input view is associated with a predicted depth and offset. Consequently, the network only obtains a shortcut representation of the rendered images. When we filter out the 50\% lowest opacity points, most of the background points in the input view are removed, resulting in a waste of Gaussian points.
In contrast, our method ensures that all Gaussians contribute to the 3D object. The geometry of our objects is nearly accurate, and removing low opacity points does not compromise the overall 3D structure. This highlights the meaningfulness and coherence of our approach in capturing the true 3D representation of objects.

\paragraph{Quantitative comparison after excluding neighboring views} \label{app: remove_near}
\begin{table}
  \caption{Novel Views for removing the 8 nearest views to input view for ShapeNet SRN dataset}
  \label{tab: removing_near_input_view}
  \centering
  \begin{tabular}{llll}
    \toprule
    \multicolumn{1}{c}{Method} & \multicolumn{3}{c}{Chairs}                \\
      & PSNR $\uparrow$     & SSIM $\uparrow$ & LPIPS $\downarrow$  \\
    \midrule
    Splatter Image  & 24.22 & 0.930 & 0.069  \\
    \midrule
    Ours & \textbf{25.29} & \textbf{0.937} & \textbf{0.067}   \\
    \bottomrule
  \end{tabular}
\end{table}
To prove that our method is not a shortcut for the input view, we conducted an experiment where we removed the views that were very close to the input view. Specifically, we excluded the nearest 8 views (index 60 to 68), as compared to Splatter Image. The results of this experiment, presented in Table \cref{tab: removing_near_input_view}, clearly indicate the superiority of our method over Splatter Image.

Our method exhibits a substantial improvement in performance compared to Splatter Image .When comparing \cref{tab: novel_view_results} and \cref{tab: removing_near_input_view}, our model shows minimal decrease in metrics when removing the values of the 8 views near the input view. However, the Splatter Image dataset exhibits a notable decrement in performance. This comparison provides evidence that our method is not simply a shortcut around the input view.
 By removing the values of the 8 views near the input view, our method decrease not much, which highlight the effectiveness of our method in enhancing the quality of the generated views when compared to Splatter Image. These results validate that our approach performs better, particularly when it comes to generating novel views that far from input views. \\
 
\paragraph{More ablation study} \label{app: ablation_param}

\begin{table}[htp]
\small
  \caption{Ablation study training on Objaverse and testing on GSO.}
  \label{tab: ablation_study_enc_dec}
  \centering
  \begin{tabular}{llll}
    \toprule
    \multicolumn{1}{c}{Method} & \multicolumn{3}{c}{Chairs}                  \\
      & PSNR $\uparrow$     & SSIM $\uparrow$ & LPIPS $\downarrow$ \\
    \midrule
    DINOv2 Encoder & 19.55 &	0.829 &	0.258 \\
    UNet Encoder &	21.06 &  0.879 & 0.111 \\
    \midrule
    Full model & \textbf{21.74} & \textbf{0.884} & \textbf{0.105}   \\
    \bottomrule
  \end{tabular}
\end{table}



We give abalation study for some extremely important hyperparameters, including the number of decoder layers and the number of Gaussians utilized to represent an object in the section. We discuss the trade-off of performance and efficiency for different hyperparameter selection.\\

\begin{figure*}
    \centering
    \includegraphics[width=0.9\textwidth]{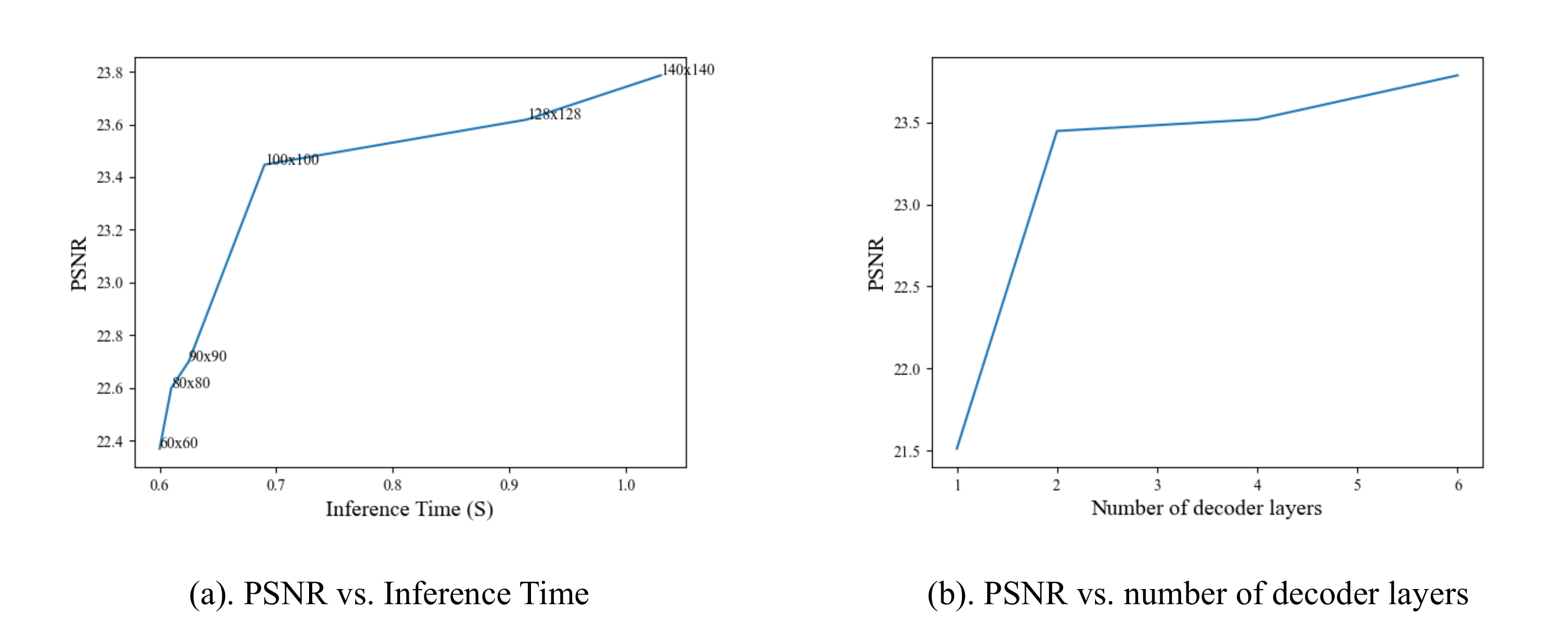}
    \caption{Ablation study for different hyperparamter selection on ShapeNet SRN chair dataset at 100k steps. (a). Inference time cost and PSNR for different number of Gaussians. We annotate the number of Gaussians at each point in the figure (e.g. $100\times 100$). (b). PSNR for different number of decoder layers.}
    \label{fig: ablation_param}
\end{figure*}
To analyze the impact of the number of Gaussians used for object representation, we conducted an ablation study. Intuitively, a higher number of Gaussians tends to enhance performance but at the cost of slower speed. Thus, we aimed to find a trade-off between model complexity and performance. The results of this study are presented in \cref{tab: ablation_study_num_Gs}, indicating that increasing the number of Gaussians leads to improved performance. However, it is crucial to note that using an excessively large number of Gaussians significantly increases the training cost.
To provide a more explicit comparison between inference time and PSNR for different numbers of Gaussians, we present \cref{fig: ablation_param} (a). When the number of Gaussians smaller 10,000, the PSNR increases significantly. However, when the number of Gaussians exceeds 10,000, the additional benefits diminish while incurring substantial time consumption. Consequently, to strike a balance between speed and quality, we ultimately opted for 10,000 Gaussians to represent an object. This choice allows us to achieve satisfactory results while maintaining reasonable training efficiency.\\

\begin{table*}[htp]
\footnotesize
  \caption{Ablation study for number of Gaussians per object on ShapeNet SRN Chair dataset for 100k step training. 3D: 3d Reconstruction speed; R: rendering speed. Test: the speed from image to all 250 novel views for each object in the test dataset.}
  \label{tab: ablation_study_num_Gs}
  \centering
  \begin{tabular}{lllllll}
    \toprule
    \multicolumn{1}{c}{Number of Gaussians}  & \multicolumn{6}{c}{Chairs}                  \\
     & 3D (S) $\downarrow$ & R (S) $\downarrow$ & Test (S) $\downarrow$ & PSNR $\uparrow$     & SSIM $\uparrow$ & LPIPS $\downarrow$ \\
    \midrule
    $128 \times 128$  & 0.24 & 0.0027 & 0.915 & \textbf{23.453} & \textbf{0.92}
& \textbf{0.11}  \\
    $100 \times 100$ & 0.14 & 0.0022 & 0.690 & 23.447 & \textbf{0.92} & \textbf{0.11}   \\
    $80 \times 80$ & 0.11 & 0.0020 & 0.610 & 22.600  & 0.91 & 0.10 \\
    \bottomrule
  \end{tabular}
\end{table*}


To enhance the refinement process in updating the 3D Gaussians, we employ multi-layer refinement in the decoder. In order to evaluate the influence of the number of layers in the decoder, we conducted an ablation study. The results, depicted in \cref{fig: ablation_param} (b), reveal that utilizing 2 layers yields significantly superior performance compared to using only 1 layer. Upon increasing the number of layers to 4 and 6, the observed improvement in performance is not as substantial as the transition from 1 layer to 2 layers. However, it is worth noting that the computational speed decreases significantly with the addition of more layers. Therefore, in most of our experiments, we opted to use 2 layers.
Selecting 2 layers strikes a balance between performance and computational efficiency. It provides a satisfactory level of accuracy while maintaining a reasonable speed for practical applications. However, in scenarios where higher accuracy is deemed necessary, the option to utilize more layers is available. This flexibility allows us to adapt the model's depth to specific requirements and trade-offs between accuracy and computational resources. This finding demonstrates the importance of incorporating multiple layers in the refinement process, highlighting the effectiveness of our approach in improving the quality of the reconstructed 3D structures. 

\paragraph{More comparison} 
The comparison between Triplane-Gaussian, Tripo \citep{TripoSR} and our methods is shown in \cref{tab: triplane_results}. Triplane Gaussian requires 3D supervision and takes longer inference time while get worse performance comparing to our model. We test on the given light-weight checkpoint in the github. Our model surpass the previous methods on both the performance and the inference speed.

\begin{table*}[htp]
\footnotesize
  \caption{Quantitative results trained on Objaverse LVIS and tested on GSO. 3D sup. means need 3D supervision.}
  \label{tab: triplane_results}
  \centering
  \begin{tabular}{lllllll}
    \toprule
       Method & PSNR $\uparrow$     & SSIM $\uparrow$ & LPIPS $\downarrow$ & 3D sup. & Inference time & Rendering time\\
    \midrule
    Triplane-Gaussian \citep{triplane-gs} & 18.61 & 0.853 & 0.159 & \Checkmark & 1.906 & \textbf{0.0025} \\
    TripoSR \citep{TripoSR} & 20.00 & 0.872 & 0.149 & \XSolidBrush & 3.291 & 22.7312 \\
    \midrule
    Ours & \textbf{23.45} & \textbf{0.897} & \textbf{0.093} &\XSolidBrush & \textbf{0.476}\ & \textbf{0.0025}\\
    \bottomrule
  \end{tabular}
\end{table*}


\end{document}